\documentclass[lettersize,journal]{IEEEtran}
\usepackage{amsmath,amsfonts}
\usepackage{algorithmic}
\usepackage{algorithm}
\usepackage{array}
\usepackage[caption=false,font=normalsize,labelfont=sf,textfont=sf]{subfig}
\usepackage{textcomp}
\usepackage{stfloats}
\usepackage{url}
\usepackage{verbatim}
\usepackage{graphicx}
\usepackage{cite}
\usepackage{multirow}
\usepackage{makecell}
\usepackage{booktabs}
\usepackage{hyperref}
\usepackage{cleveref}
\usepackage{bm}
\Crefname{figure}{Fig.}{Figs.}
\begin{document}

\title{DMSSN: Distilled Mixed Spectral-Spatial Network for Hyperspectral Salient Object Detection}

\author{Haolin Qin, Tingfa Xu$^{\dagger}$, Peifu Liu, Jingxuan Xu, Jianan Li$^{\dagger}$
\thanks{H. Qin, T. Xu, P. Liu, and J. Li are with Beijing Institute of Technology, 100081 Beijing, China. E-mail:\{3120225333, ciom\_xtf1, 3120210580, lijianan\}@bit.edu.cn.}
\thanks{J. Xu is with Peter the Great St.Petersburg Polytechnic University, 195257 St. Petersburg, Russia. E-mail: 1735200206@qq.com.}
\thanks{J. Li and T. Xu are also with the Key Laboratory of Photoelectronic Imaging Technology and System, Ministry of Education of China, Beijing 100081, China.}
\thanks{T. Xu is also with Chongqing Innovation Center, Beijing Institute of Technology, Chongqing 401135, China.}
\thanks{$^{\dagger}$ Correspondence to: Tingfa Xu and Jianan Li.}
}

\markboth{Journal of \LaTeX\ Class Files,~Vol.~14, No.~8, August~2021}%
{Qin \MakeLowercase{\textit{et al.}}: DMSSN: Distilled Mixed Spectral-Spatial Network for Hyperspectral Salient Object Detection}

\maketitle

\begin{abstract}
Hyperspectral salient object detection (HSOD) has exhibited remarkable promise across various applications, particularly in intricate scenarios where conventional RGB-based approaches fall short. Despite the considerable progress in HSOD method advancements, two critical challenges require immediate attention. Firstly, existing hyperspectral data dimension reduction techniques incur a loss of spectral information, which adversely affects detection accuracy. Secondly, previous methods insufficiently harness the inherent distinctive attributes of hyperspectral images (HSIs) during the feature extraction process. To address these challenges, we propose a novel approach termed the Distilled Mixed Spectral-Spatial Network (DMSSN), comprising a Distilled Spectral Encoding process and a Mixed Spectral-Spatial Transformer (MSST) feature extraction network. The encoding process utilizes knowledge distillation to construct a lightweight autoencoder for dimension reduction, striking a balance between robust encoding capabilities and low computational costs. The MSST extracts spectral-spatial features through multiple attention head groups, collaboratively enhancing its resistance to intricate scenarios. Moreover, we have created a large-scale HSOD dataset, HSOD-BIT, to tackle the issue of data scarcity in this field and meet the fundamental data requirements of deep network training. Extensive experiments demonstrate that our proposed DMSSN achieves state-of-the-art performance on multiple datasets. We will soon make the code and dataset publicly available on https://github.com/anonymous0519/HSOD-BIT.
\end{abstract}


\section{Introduction}
\begin{figure}[t]
  \centering
  \includegraphics[width=1\linewidth]{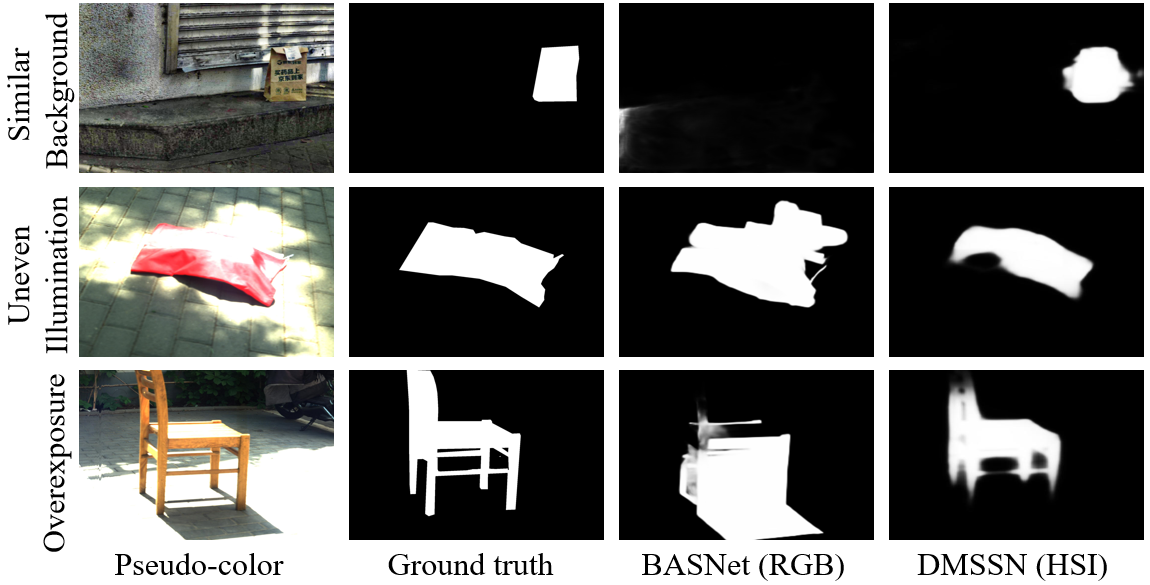}
  \caption{Comparison of the predicted results from RGB-based and HSI-based methods under challenging conditions, such as a similar background, uneven illumination, and overexposure, HSI-based methods prioritize the importance of spectral salience, leading to more dependable detection results.}
  \label{fig:motivation}
\end{figure}

\IEEEPARstart{H}{yperspectral} image processing is one of the key technologies in remote sensing \cite{li2023lightweight}, and has emerged in various fields \cite{nasrabadi2013hyperspectral, gao2019optical}. As one application of this technology, hyperspectral salient object detection (HSOD), whose primary objective is to distinguish salient objects from complex backgrounds, plays a critical role in remote sensing tasks \cite{wang2022multiscale, zheng2023boundary}. Previous methods typically adopt RGB-based methods to process hyperspectral data through spectral compression. However, they discard the majority of spectral information\cite{li2019distortion}, leading to poor performance in challenging conditions. The conventional RGB-based methods primarily focus on spatial features, hindering the ability to discriminate between objects and backgrounds when their colors are similar. As illustrated in \Cref{fig:motivation}, RGB-based methods predict distorted results when faced with similar backgrounds, uneven illumination, or overexposure. 

\begin{figure}[t]
  \centering
  \includegraphics[width=1\linewidth]{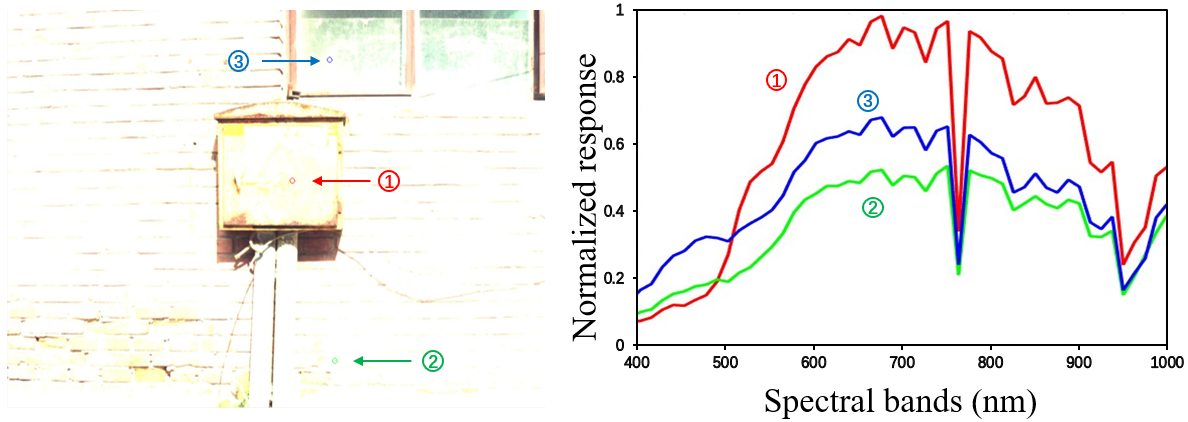}
  \caption{Under overexposure, the foreground objects and background may appear similar in color, but their spectral curves exhibit significant differences, enabling their distinguishability.}
  \label{fig:curves}
\end{figure}

In contrast to traditional RGB images, hyperspectral images (HSIs) record object properties using spectral curves rather than only three primary colors. This unique characteristic endows HSIs with a wealth of spectral information, enabling more precise and accurate representation of objects \cite{wu2022hyperspectral}. As depicted in \Cref{fig:curves}, when subjected to overexposure lighting conditions, RGB images struggle to distinguish between similar colors in the foreground and background. However, spectral curves exhibit substantial variations even among objects with similar colors, as they primarily reflect material properties and remain unaffected by complex conditions. Therefore, focusing on spectral salience will be an effective solution to improve HSOD performance with complex background and lighting conditions perturbations.

Recently, some studies have proposed HSOD methods exploiting spectral information, but with limited success. Among them, the traditional methods \cite{zhang2018salient} utilized mathematical algorithms to calculate the saliency score of each pixel. They heavily relied on handcrafted shallow features and prior knowledge, leading to low robustness, limited ability to represent object salience, and suboptimal detection performance in complex conditions. Recognizing these shortcomings, we considered introducing deep learning techniques, which have achieved great success in multiple computer vision fields. Through extensive experiments and analysis, we found that two challenges limit the further development of this technology in HSOD tasks.

The first challenge is the redundancy of excessive spectral dimensions in hyperspectral images, which brings high computational costs \cite{cai2021hypergraph}. Consequently, data dimension reduction emerges as a crucial and indispensable step in the HSI-based method. Most existing HSOD methods adopt mathematical algorithms, such as Principal Component Analysis (PCA), to reduce data dimension \cite{yu2022unsupervised}. However, these traditional algorithms are constrained by linear transformations, often resulting in the loss of valuable information and a subsequent decrease in detection accuracy. As a result, there arises a need for more sophisticated techniques capable of overcoming these limitations and preserving essential spectral information, thereby enhancing the performance of HSOD methods.

To address the aforementioned limitations, we propose a novel method termed the Distilled Spectral Encoding process, which leverages an autoencoder with a knowledge distillation strategy. Specifically, we introduce a spectral homogenization process based on the Gaussian Mixture Model (GMM) as a preprocessing step, which helps to reduce background complexity and enhances object contrast. The homogenized output is subsequently fed into an autoencoder, capable of complex nonlinear transformations and endowed with superior data encoding capacity compared to previous algorithms \cite{zhou2019learning}, all while minimizing information loss. Notably, deepening the autoencoder network can enhance its encoding ability, but also leads to increased computational costs. To address this issue, we employ a lightweight student autoencoder trained under the guidance of a deep teacher autoencoder through the knowledge distillation strategy. This strategy produces enhanced dimension reduction results without incurring additional computational costs. 

The second challenge revolves around the concurrent utilization of rich spectral and spatial information in hyperspectral images to enhance detection performance. Existing methods frequently concentrate solely on a single category of information or independently employ two categories of features without facilitating meaningful interaction between them.

To overcome the above issue, we introduce a feature extraction backbone network termed the Mixed Spectral-Spatial Transformer (MSST), meticulously designed to leverage the inherent characteristics of HSIs. The MSST comprises a series of stacked Mixed Spectral-Spatial (MSS) blocks. To enhance the ability to simultaneously focus on both spectral and spatial information, the MSS block employs a divide-and-conquer strategy for feature extraction. It incorporates a mixed spectral-spatial attention mechanism by utilizing distinct attention head groups, with each group dedicated to extracting one category of features. Subsequently, the extracted features are merged, facilitating per-category feature learning and cross-category feature interaction. This design renders the proposed MSST well-suited for harnessing spectral-spatial information in the context of hyperspectral image processing.

Building upon the Distilled Spectral Encoding process and the MSST network, we present the Distilled Mixed Spectral-Spatial Network (DMSSN). The DMSSN compresses the input hyperspectral image data along the spectral dimension and then extracts both spectral and spatial features concurrently using the reduced data. Based on the learned features, DMSSN produces highly accurate saliency maps for a hyperspectral image while ensuring high efficiency. These innovative components empower DMSSN to deliver more efficient and accurate HSOD performance. By amalgamating the benefits of the Distilled Spectral Encoding process and the MSST network, DMSSN effectively addresses challenges in HSOD tasks. As a result, our proposed method offers a compelling solution for enhancing the performance of hyperspectral image processing, demonstrating remarkable potential in various practical applications.

Deep learning techniques heavily rely on substantial labeled datasets to train deep methods effectively. However, for the HSOD task, the availability of large-scale datasets is currently limited, impeding research progress in this domain. To solve this problem, we create HSOD-BIT, a novel large-scale HSOD dataset offering two significant advantages over existing datasets \cite{imamoglu2018hyperspectral}. Firstly, HSOD-BIT boasts a larger data volume, comprising a total of $319$ hyperspectral images. These images encompass $200$ visible to near-infrared bands and a high spatial resolution of $1240 \times 1680$ pixels, thereby fulfilling the fundamental data requirements for training deep methods. Secondly, HSOD-BIT incorporates challenging data with multiple complex conditions, including similar backgrounds, uneven illumination, and overexposure, in addition to conventional scenes. This diversity ensures that the dataset can effectively evaluate the performance of HSOD methods under various real-world scenarios. 

We firmly believe that the introduction of HSOD-BIT will serve as a substantial contribution to the advancement of HSOD research. The dataset's substantial volume and diverse challenges offer ample opportunities for researchers to develop and fine-tune more robust and accurate HSOD methods. As a result, we envision that HSOD-BIT will be instrumental in driving progress and innovation in the field of HSOD.

The experimental results robustly affirm the effectiveness of our proposed DMSSN, showcasing its state-of-the-art performance on the collected HSOD-BIT dataset. Our method surpasses several popular RGB-based methods, particularly excelling in complex scenarios. Moreover, we extend the proposed MSST backbone network to remote sensing vision tasks. In remote sensing target detection and classification tasks, the MSST-based method exhibits outstanding performance, further emphasizing the versatility and efficacy of our approach across diverse applications within the hyperspectral image processing domain.

To sum up, this work makes the following contributions:
\begin{itemize}
\item We propose a novel approach, termed Distilled Spectral Encoding, which employs an autoencoder with a knowledge distillation strategy to achieve efficient spectral dimension reduction.
\item We propose the Mixed Spectral-Spatial Transformer, a novel backbone network that is specifically designed for hyperspectral image processing. 
\item We have curated a new dataset, called HSOD-BIT, which features a large-scale collection of hyperspectral images with high spatial resolution, broad spectral range, and challenging scenes.
\end{itemize} 

\section{Related Work}

\subsection{Hyperspectral Salient Object Detection}
Since the pioneering work of Itti \textit{et al.} \cite{itti1998model}, the field of salient object detection (SOD) has witnessed a surge of RGB-based methods \cite{zhao2023learning}. However, these methods struggle in challenging conditions such as similar backgrounds, uneven illumination, and over-exposure, as they solely rely on the three primary colors. Such scenarios underscore the intrinsic limitations of RGB-based techniques in distinguishing salient objects from complex backgrounds. Recognizing these limitations, Liang \textit{et al.} \cite{liang2013salient} proposed the first HSIs-based method for the HSOD task by substituting trichromatic spectrum for color components. In the wake of this innovation, several traditional HSOD methods emerged\cite{zhang2018salient}, leveraging shallow features such as spectral contrast, spectral angular distance, and spectral gradient. These features, derived from the unique spectral properties of HSIs, offered new avenues for detecting salient objects. However, their effectiveness was limited by the requirement for prior knowledge in manual feature design, and they often proved inadequate in fully capturing the complexity of object saliency in diverse scenes.

To overcome these limitations, several deep learning-based methods demonstrated impressive performance in remote sensing image based HSOD tasks \cite{gu2023orsi, li2019nested, li2021multi}. For instance, Imamoglu \textit{et al.} \cite{imamouglu2019salient} introduced an unsupervised convolutional neural network (CNN) tailored to the lack of hyperspectral data. Nonetheless, this approach exhibited limitations, including susceptibility to noise, compromised generalization capabilities, and limited applicability in practical scenarios. Similarly, Huang \textit{et al.} \cite{huang2021salient} proposed a wireless network with saliency optimizations. This method suffered from a lack of interactive between spatial and spectral information and the incorporation of the non-differentiable SLIC algorithm for superpixel generation, which hinders seamless end-to-end training. These methods fail to fully exploit the rich hyperspectral characteristics during feature extraction, resulting in lost opportunities for maximizing detection performance. Therefore, the objective of this work is to enhance the detection performance of HSOD methods specifically for natural scenes.

\subsection{Transformer for Hyperspectral Image Processing}
Hyperspectral images inherently encompass both spatial and spectral information \cite{yang2021cross}, necessitating the integration of both aspects in hyperspectral image processing. Despite advancements, existing methods exhibit constraints in harnessing this integrative potential. For instance, some methods utilize parallel or convolutional structures for feature extraction. While parallel structures are adept at isolating distinct feature categories, they come with the drawback of high computational demands \cite{huang2021salient}. On the other hand, convolutional architectures, though beneficial for local feature analysis, might not fully encompass the global context intrinsic to hyperspectral data \cite{roy2019hybridsn}.

Transformers, renowned for their adeptness at modeling long-range dependencies \cite{dosovitskiy2020image}, have revolutionized computer vision, aligning well with hyperspectral imagery's intrinsic characteristics. This alignment has propelled their adoption for a variety of hyperspectral image processing tasks, such as classification, segmentation, and reconstruction \cite{roy2023spectral, kang2023self, qin2023factorization}, showcasing substantial advancements. However, the conventional use of linear projections within these Transformer models can inadvertently diminish vital local spectral and spatial nuances. Furthermore, the computational intensity associated with Transformer architectures poses a challenge for their application in scenarios requiring scalability and timeliness \cite{he2021spatial}. To navigate these hurdles, we introduce the innovative Mixed Spectral-Spatial Transformer (MSST). The MSST is meticulously designed to synergize spectral and spatial features, aiming to transcend current performance benchmarks while optimizing computational efficiency in hyperspectral image processing.

\subsection{HSOD Datasets}
The scarcity of data available for the HSOD task has been a persistent challenge \cite{cao2015salient}. While some researchers have attempted to collect hyperspectral images from public resources on the Internet \cite{liang2013salient, yan2016salient}, the quantity and quality of these images are insufficient to support network training. This is primarily due to the lack of pixel-level annotations and the limited number of images available in these datasets. Furthermore, some works have focused on remote sensing images \cite{liu2022global, wang2022hybrid}, but the differences between remote sensing images and natural scene images limit the transfer learning of methods. Recently, Imamoglu \textit{et al.} \cite{imamoglu2018hyperspectral} collected a small-scale HSOD dataset consisting of only 60 hyperspectral images. However, this dataset is inadequate for deep learning-based method training, especially for supervised learning that requires large amounts of data. Additionally, previous datasets have only collected hyperspectral data from conventional scenes, neglecting diverse environmental conditions. 

Therefore, it is worthwhile to consider the collection of a large-scale dataset that takes into account diverse environmental conditions to fully exploit the valuable spectral information present in hyperspectral images. Addressing the issue of data scarcity and diversifying the dataset will play a pivotal role in advancing the development and generalization of HSOD methods. By curating a large-scale dataset that captures various environmental scenarios, we can significantly enhance the training process and enable HSOD methods to perform more effectively in real-world applications involving natural scene images.

\section{Method}

\begin{figure*}[t]
  \centering
   \includegraphics[width=1\linewidth]{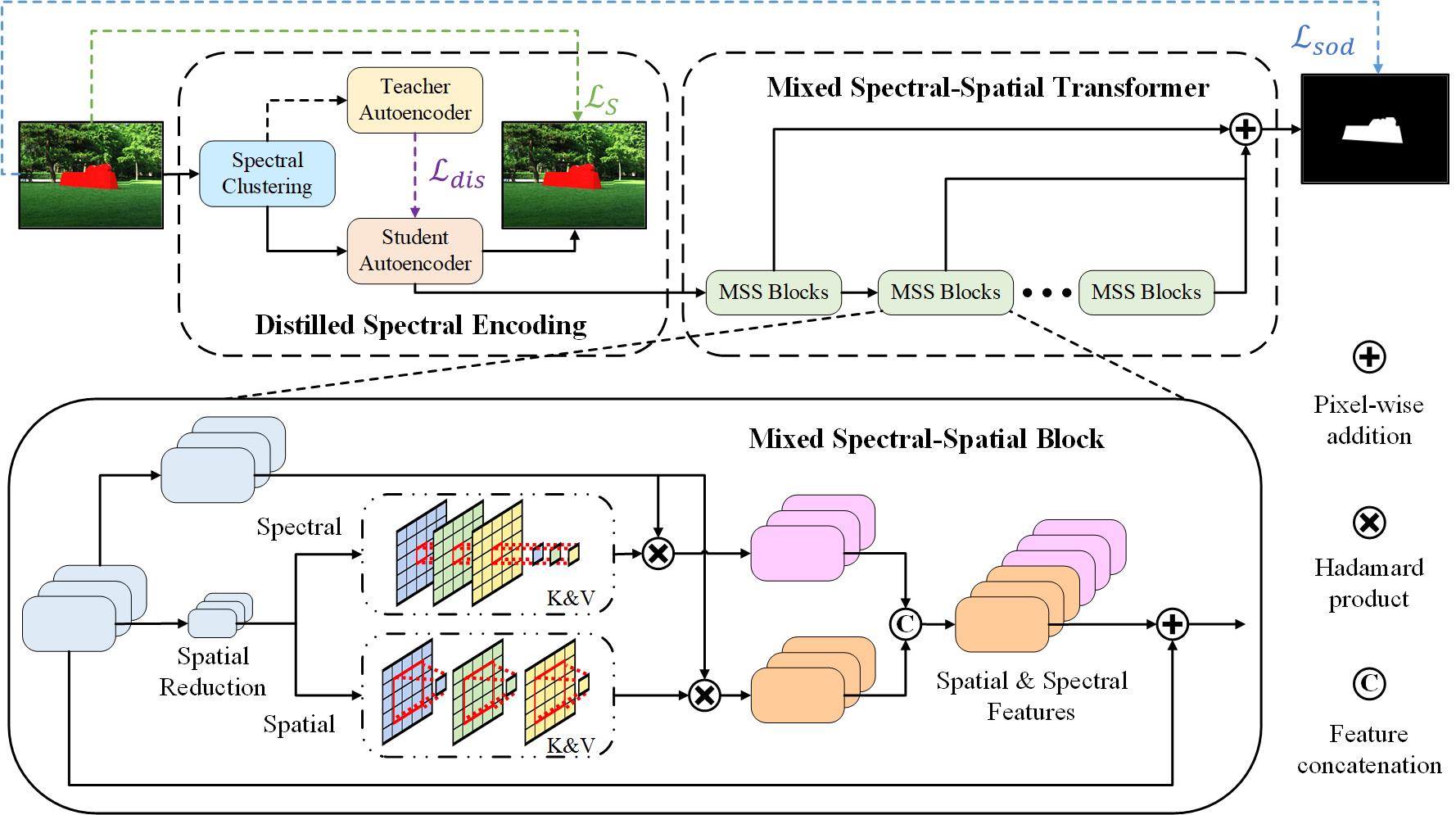}
   \caption{The overall architecture of the proposed Distilled Mixed Spectral-Spatial Network (DMSSN) is shown in the top part of the figure. The bottom part illustrates the Mixed Spectral-Spatial Transformer (MSST) block, where queries (Q) are obtained through linear projection, and Keys(K) and values(V) are generated through dual branches with spectral and spatial features respectively. The circle symbols with sum, cross and C represent pixel-wise addition, Hadamard product and feature concatenation respectively.}
   \label{fig:Overview}
\end{figure*}

The hyperspectral salient object detection (HSOD) task aims to generate a saliency map ${\textbf{\textit{Y}}} \in \mathbb{R}^{\rm{H \times W \times 1}}$ that labels object and background pixels in a hyperspectral image ${\textbf{\textit{I}}} \in \mathbb{R}^{\rm{H \times W \times C}}$, where $\rm{H}$, $\rm{W}$, and $\rm{C}$ represent height, width, and spectral dimension, respectively \cite{ren2020salient}. To tackle the HSOD task, we propose the Distilled Mixed Spectral-Spatial Network (DMSSN), as illustrated in \Cref{fig:Overview}. DMSSN consists of two primary components:

(i) \textbf{The Distilled Spectral Encoding process:} This component employs spectral homogenization as a preprocessing step and knowledge distillation to construct an autoencoder that efficiently performs spectral dimension reduction. This process enhances computational efficiency while retaining essential spectral information. 

(ii) \textbf{The Mixed Spectral-Spatial Transformer (MSST):} This feature extraction backbone network learns spectral and spatial features from the encoded output through a series of stacked Mixed Spectral-Spatial blocks and outputs the final saliency map. 

By combining the strengths of these two components, DMSSN can efficiently and accurately detect salient objects in hyperspectral images. This architecture allows for comprehensive feature learning and facilitates the effective utilization of both spectral and spatial information, resulting in improved performance and robustness in HSOD tasks.

\subsection{Distilled Spectral Encoding}
Due to the high number of spectral dimensions present in hyperspectral images, directly using them in the feature extraction process would require a significant amount of computation. To overcome this challenge, we propose a knowledge distillation strategy that employs a lightweight autoencoder to efficiently reduce spectral dimensionality without sacrificing crucial information. 

Specifically, we initiate a preprocessing step of spectral homogenization, which enhances the contrast between objects and backgrounds in the hyperspectral image. The resulting homogenized output serves as input to the autoencoder, which leverages nonlinear transformations to perform high-quality data encoding. To strike a balance between accuracy and efficiency, we adopt a knowledge distillation approach, training a distilled student autoencoder under the guidance of a teacher autoencoder. This approach allows us to generate improved dimension reduction results without incurring additional computational costs.

\subsubsection{Spectral Homogenization}\label{subsubsec3.1.1}
Due to the influence of environmental factors, the spectral curve of objects will fluctuate \cite{liang2018material}, resulting in increased complexity of the background and reduced salience of target objects. To address this issue, we employ a Spectral Homogenization process using the Gaussian Mixture Model (GMM) \cite{zong2018deep} as a preprocessing step to highlight salient objects. Suppose that the input hyperspectral image ${\textbf{\textit{I}}}$ records $\rm{M}$ materials, each of which corresponds to a unique spectral curve. We can approximate the probability distribution of the material to which each pixel belongs using $\rm{M}$ mixed multivariate Gaussian distribution functions. The probability density function of the GMM is given by:
\begin{equation}
\emph{\textbf{P}}(\textbf{\textit{I}}) = \sum_{\emph{i}=1}^{\rm{M}}\alpha_{\emph{i}}\emph{\textbf{p}}({\textbf{\textit{I}}}|\mu_{\emph{i}},\sigma_{\emph{i}}), \label{eq1}
\end{equation}
where $\emph{\textbf{P}}(\cdot)$ denotes the probability density function, $\alpha_i$, $\mu_i$, $\sigma_i$, and $\emph{\textbf{p}}(\cdot)$ correspond to the weight, expectation, standard deviation, and probability distribution, respectively, of the $i$-th sub-model. The value of $\rm{M}$ is usually set manually, defaulting to $50$, while the remaining parameters are estimated through the Expectation Maximization (EM) algorithm.

The GMM algorithm calculates the likelihood of a pixel belonging to each material based on its spectral curve. We then assign each pixel to the material with the highest likelihood and replace the original spectral curve with the average curve for that material. As a result, the generated hyperspectral data, denoted as ${\textbf{\textit{G}}} \in \mathbb{R}^{\rm{H \times W \times C}}$. Specifically, the vector corresponding to a certain spatial position can be calculated as follows:
\begin{equation}
{\textbf{\textit{G}}}_{\emph{i,j}}^{\emph{m}} = \frac{1}{\rm{N}_{\emph{m}}}\sum_{n=1}^{\rm{N}_{\emph{m}}}{\textbf{\textit{I}}}_{\emph{n}}^{\emph{m}} \in \mathbb{R}^{1 \times 1 \times \rm{C}}.  \label{eq2}
\end{equation}
Where $i=1,\dots,\rm{W}$ and $j=1,\dots,\rm{H}$ represent the spatial coordinate position. $m=1,\dots,\rm{M}$ represents the category with the highest probability of the vector among the M materials. ${\textbf{\textit{G}}}_{\emph{i,j}}^{\emph{m}}$ denotes the spectral curve of the vector, which is labeled the $m$-th material, at the spatial coordinate $(i, j)$ in the hyperspectral image. ${\rm{N}_{\emph{m}}}$ represents the total number of vectors labeled $m$. ${\textbf{\textit{I}}}_{\emph{n}}^{\emph{m}} \in \mathbb{R}^{1 \times 1 \times \rm{C}}$ denotes the spectral curve of the $n$-th one in the vectors belonging to the $m$-th material. 

Following the spectral homogenization process, the spectral curves of objects composed of the same material but located in different regions become identical. These operation enhances the contrast between different materials and facilitates their detection. By grouping pixels with similar spectral characteristics, the homogenized process helps in emphasizing the distinctions between various materials present in the hyperspectral image. Consequently, the subsequent hyperspectral salient object detection process benefits from the enhanced contrast and improved discriminate information, leading to more accurate and effective identification of salient objects in complex natural scenes.

\subsubsection{Knowledge Distillation Strategy}
We propose an autoencoder framework to reduce the spectral dimensionality of data while preserving relevant information. This framework consists of two key stages: the encoding stage and the decoding stage. In the encoding stage, the autoencoder generates an encoded feature map from the input hyperspectral data. This feature map represents a compressed representation of the original data, capturing its essential features and reducing its spectral dimensionality. Subsequently, in the decoding stage, the autoencoder reconstructs the original hyperspectral data from the encoded feature map. This reconstruction process aims to approximate the original data as closely as possible, ensuring that relevant information is retained and the quality of the reconstructed data is preserved. 

Let ${\textbf{\textit{G}}} \in \mathbb{R}^{\rm{H \times W \times C}}$ represent the input spectral homogenization result. The encoding process is denoted as $\boldsymbol{\phi}(\cdot)$ and can be mathematically expressed as:
\begin{equation}
{\textbf{\textit{E}}} = \boldsymbol{\phi}({\textbf{\textit{G}}}) \in \mathbb{R}^{\rm{H \times W \times C'}}. \label{eq3}
\end{equation}
In the encoding process, we preserve the spatial resolution of the input while reducing the spectral dimensions, such that ${\rm{C'} < \rm{C}}$. The resulting encoded feature map, denoted as ${\textbf{\textit{E}}}$, can be considered as the outcome of the dimension reduction process.
The decoding process, denoted as $\boldsymbol{\psi}(\cdot)$, is defined as follows:
\begin{equation}
{\textbf{\textit{D}}} = \boldsymbol{\psi}({\textbf{\textit{E}}}) \in \mathbb{R}^{\rm{H \times W \times C}}. \label{eq4}
\end{equation}
The decoding results ${\textbf{\textit{D}}}$ are employed to facilitate model convergence, as further elaborated in subsequent sections.

The depth of an autoencoder significantly impacts its encoding capability, suggesting that to achieve high-quality data dimension reduction results, a deep encoder-decoder network is often preferred. However, when dealing with high-resolution hyperspectral images as input, the computational and memory costs associated with a deep autoencoder become a concern. The computational expense of a deep autoencoder can hinder the development and implementation of subsequent feature extraction networks. 

To overcome this challenge and strike a balance between encoding capability and computational efficiency, we propose a knowledge distillation strategy. This approach allows us to construct a lightweight autoencoder that retains significant encoding ability while reducing both computational and memory costs. By distilling knowledge from a deep teacher autoencoder to guide the training of a more compact student autoencoder, we achieve an efficient yet effective dimension reduction framework for hyperspectral images.

\begin{figure}[t]
  \centering
   \includegraphics[width=1\linewidth]{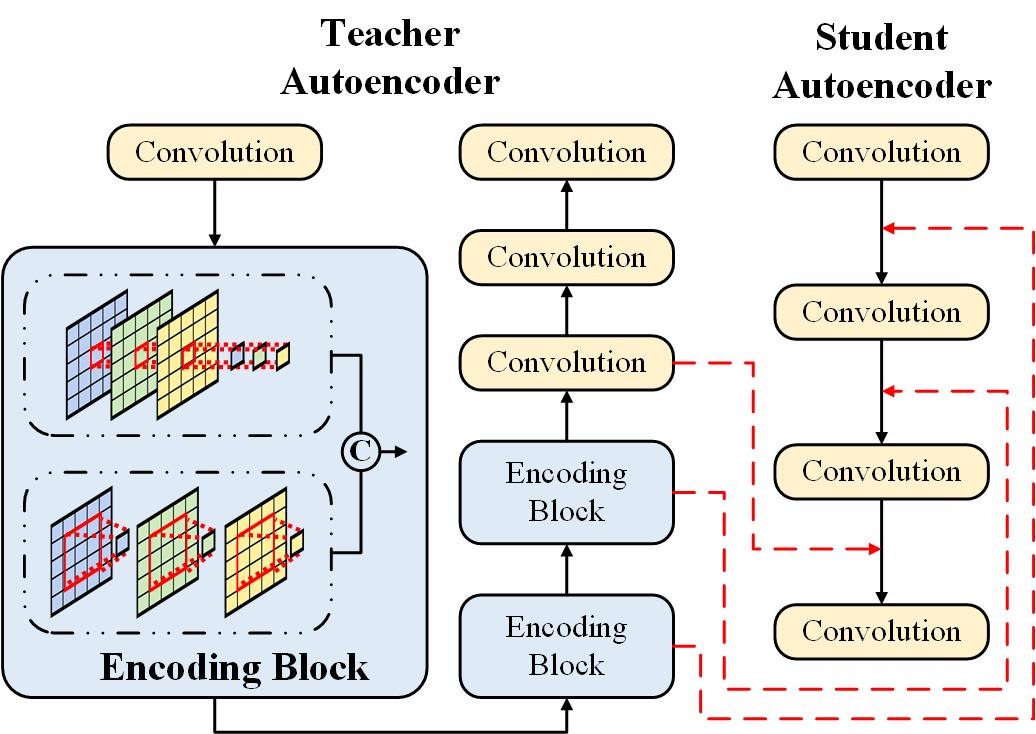}
   \caption{Network architecture for autoencoders. The left part illustrates the teacher autoencoder, which consists of encoding blocks and convolutional layers. The encoding block implements simultaneous extraction and fusion of spatial and spectral features. The right part provides the details of the student autoencoder, composed only of convolutional layers. The red dashed line linking the autoencoders indicates the distillation process.}
   \label{fig:auto}
\end{figure}

\textbf{Teacher Autoencoder.}
We construct a deep autoencoder to function as the teacher autoencoder, as illustrated in \Cref{fig:auto}. The teacher autoencoder employs dual-branch encoding blocks during its encoding stage. One branch focuses on spatial information by splitting the feature maps along the spectral dimension, while the other branch traverses all pixels to capture spectral information comprehensively. The concatenation of features from these branches aims to facilitate effective information interaction. The operations within each branch closely follow those detailed in \Cref{subsec3.2} and can be mathematically expressed as:
\begin{equation}
{\textbf{\textit{E}}}_{\rm{T}}' = \boldsymbol{\varphi}(\boldsymbol{\delta}{({\textbf{\textit{F}}}_{\rm{spe}},{\textbf{\textit{F}}}_{\rm{spa}}))}, \label{eq5}
\end{equation}
where ${\textbf{\textit{E}}}_{\rm{T}}' \in \mathbb{R}^{\rm{H \times W \times C'}}$ denotes the output of the encoding block in the teacher autoencoder. ${\textbf{\textit{F}}}_{\rm{spe}} \in \mathbb{R}^{\rm{H \times W \times C}}$ and ${\textbf{\textit{F}}}_{\rm{spa}} \in \mathbb{R}^{\rm{H \times W \times C}}$ represent the spectral and spatial features extracted by different branches. $\boldsymbol{\delta}(\cdot)$ represents the concatenation operation and $\boldsymbol{\varphi}(\cdot)$ represents a linear layer. 
The linear layers are employed to modify the spectral dimension during the encoding and decoding stages, which generates six feature maps: $\{{\textbf{\textit{E}}}_{\rm{T}}^{1}, {\textbf{\textit{E}}}_{\rm{T}}^{2}, {\textbf{\textit{E}}}_{\rm{T}}, {\textbf{\textit{D}}}_{\rm{T}}^{2}, {\textbf{\textit{D}}}_{\rm{T}}^{1}, {\textbf{\textit{D}}}_{\rm{T}} \}$. ${\textbf{\textit{E}}}_{\rm{T}}^{i} \in \mathbb{R}^{\rm{H \times W \times C^{\emph{i}}}}$ and ${\textbf{\textit{D}}}_{\rm{T}}^{i} \in \mathbb{R}^{\rm{H \times W \times C^{\emph{i}}}}$ denote the encoding and decoding intermediate results, where $i=1,2$ and ${\rm{C'} < \rm{C^2} < \rm{C^1} < \rm{C}}$. In addition, ${\textbf{\textit{E}}}_{\rm{T}} \in \mathbb{R}^{\rm{H \times W \times C'}}$ and ${\textbf{\textit{D}}}_{\rm{T}} \in \mathbb{R}^{\rm{H \times W \times C}}$ denote the final encoding and decoding results respectively. 

Indeed, the deep teacher autoencoder exhibits formidable encoding capability owing to its complex structure and nonlinear transformations. This empowers it to achieve superior spectral dimension reduction while retaining essential information from the original hyperspectral data. Nevertheless, the same complexity that endows it with robust encoding prowess comes with associated costs. The increased depth and complexity of the teacher autoencoder result in higher computational and memory costs during both training and inference. Processing high-resolution hyperspectral images through the deep autoencoder necessitates substantial computational resources, which can be impractical and resource-intensive for certain applications or environments.

\textbf{Student Autoencoder.}
The student autoencoder is deliberately crafted as a lightweight network, comprising only two linear layers in both the encoding and decoding stages. This intentional design choice aims to drastically reduce computational and memory costs compared to the deep teacher autoencoder. During forward propagation, the network generates feature maps: $\{ {\textbf{\textit{E}}}_{\rm{S}}^{1},{\textbf{\textit{E}}}_{\rm{S}}, {\textbf{\textit{D}}}_{\rm{S}}^{1}, {\textbf{\textit{D}}}_{\rm{S}} \}$, where ${\textbf{\textit{E}}}_{\rm{S}}^{1} \in \mathbb{R}^{\rm{H \times W \times C^2}}$ corresponds to the encoding intermediate result, ${\textbf{\textit{E}}}_{\rm{S}} \in \mathbb{R}^{\rm{H \times W \times C'}}$ corresponds to the encoding result, ${\textbf{\textit{D}}}_{\rm{S}}^{1} \in \mathbb{R}^{\rm{H \times W \times C2}}$ corresponds to the decoding intermediate result, and ${\textbf{\textit{D}}}_{\rm{S}} \in \mathbb{R}^{\rm{H \times W \times C}}$ corresponds to the decoding result.

Although the stacked linear layers significantly reduce computational and memory costs, the encoding capability is somewhat limited due to the reduced complexity of the network. Compared to the deep teacher autoencoder with its powerful encoding ability, the student autoencoder may not achieve the same level of dimension reduction quality and information preservation.

\textbf{Distillation Process.} 
The knowledge distillation strategy is implemented in two steps. Firstly, we train the teacher autoencoder, minimizing the disparity between encoding input and decoding output. The optimal parameters are saved as a pre-trained model. Secondly, we train the proposed DMSSN, utilizing the pre-trained model to guide the student autoencoder. The knowledge distillation strategy facilitates considerable encoding ability in the lightweight student autoencoder at a low cost.

Specifically, during training, three hiding layers accept knowledge from three corresponding guidance layers of the teacher autoencoder, which correspond to three groups: $\{ {\textbf{\textit{E}}}_{\rm{T}}^{2}; {\textbf{\textit{E}}}_{\rm{S}}^{1} \}$, $\{ {\textbf{\textit{E}}}_{\rm{T}}; {\textbf{\textit{E}}}_{\rm{S}} \}$, and $\{ {\textbf{\textit{D}}}_{\rm{T}}^{2}; {\textbf{\textit{D}}}_{\rm{S}}^{1} \}$. The feature maps generated by corresponding layers within each group have the same dimension, and the similarity between them is quantified by the spectral-spatial mixing distance. The mixing distance is calculated by considering both the spatial Euclidean distance and the spectral cosine similarity. For instance, the mixing distance for one group is $\emph{Dis}(\cdot,\cdot)$, which is defined as follows:
\begin{equation}
\begin{aligned}
  \bf\emph{Dis}({\textbf{\textit{E}}}_{\rm{T}},{\textbf{\textit{E}}}_{\rm{S}}) &= \frac{1}{\rm{HW}}(\Vert({\textbf{\textit{E}}}_{\rm{T}} - {\textbf{\textit{E}}}_{\rm{S}})\Vert_{\rm{F}} \\
  &- \sum_{i=1}^{\rm{HW}} \bf\emph{cos}^{-1}(\frac{ \Vert {\textbf{\textit{E}}}_{\rm{T,\emph{i}}}^{\rm{\top}}{\textbf{\textit{E}}}_{\rm{S,\emph{i}}} \Vert }{\Vert {\textbf{\textit{E}}}_{\rm{T,\emph{i}}} \Vert_{\rm{2}} \times \Vert {\textbf{\textit{E}}}_{\rm{S,\emph{i}}} \Vert_{\rm{2}}})), \label{eq6}
\end{aligned}
\end{equation}
where $\Vert \cdot \Vert_{\rm{F}}$ stands for calculating Frobenius norm. ${\textbf{\textit{E}}}_{\rm{T,\emph{i}}}$ and ${\textbf{\textit{E}}}_{\rm{S,\emph{i}}}$ represent a certain vector in the feature map, and the cosine similarity of the matrix is calculated by calculating the average value between vectors. By continuously reducing the mixing distance of each group during training, the knowledge distillation strategy transfers the knowledge learned from the deep teacher encoder to the lightweight student encoder. 

As a consequence of this knowledge transfer, the student autoencoder achieves a harmonious equilibrium between performance and computational cost. It acquires valuable encoding ability from the teacher autoencoder while maintaining a more efficient network structure. The knowledge distillation process enables the student autoencoder to achieve high-quality spectral dimension reduction and information preservation, making it an effective solution for hyperspectral image processing tasks that require both accuracy and computational efficiency.

\subsection{Mixed Spectral-Spatial Transformer}\label{subsec3.2}
\Cref{fig:Overview} illustrates the Mixed Spectral-Spatial Transformer (MSST) feature extraction backbone network, which consists of a patch embedding layer for feature mapping \cite{sun2022spectral} and multiple Mixed Spectral-Spatial (MSS) blocks arranged into several stages. Several circle symbols in \Cref{fig:Overview} are defined as follows: The circle symbol with a sum inside signifies pixel-by-pixel addition, combining two feature maps by adding their corresponding pixel values. The circle symbol with a cross inside represents the Hadamard product, indicating the point-wise multiplication of two feature maps. The circle symbol with a `C’ inside denotes concatenation along the feature dimension, allowing for the integration of different feature maps. 

The MSST takes the encoded feature map ${\textbf{\textit{E}}}_{\rm{S}}$ of the lightweight student autoencoder as input. As each stage, stacked by several MSS blocks, the MSST generates feature maps with different resolutions, leading to a series of hierarchical pyramid feature maps $\{ {\textbf{\textit{F}}}_{i} \in \mathbb{R}^{\rm{H_{\emph{i}} \times W_{\emph{i}} \times C_{\emph{i}}}} | \emph{i}=1,\dots, \rm{L}\}$, where $\rm{L}$ is the number of stages set to $4$ in practice. Finally, the MSST employs a typical Feature Pyramid Network (FPN) structure \cite{lin2017feature} to calculate the final saliency map ${\textbf{\textit{Y}}}$. The FPN is responsible for integrating information from different levels of the hierarchical pyramid feature maps to achieve comprehensive and multi-scale feature representations.

\subsubsection{Mixed Spectral-Spatial Block}
The MSS block utilizes the self-attention mechanism~\cite{wang2022pvt}, which involves extracting queries ${\textbf{\textit{Q}}}$, keys ${\textbf{\textit{K}}}$, and values ${\textbf{\textit{V}}}$. The three steps of the MSS block for extracting spectral-spatial features are as follows: i) linear projection for obtaining queries; ii) grouping attention heads to acquire keys with spectral or spatial information; and iii) spectral-spatial multi-head attention calculation.

\textbf{Obtaining Queries.}
The input feature map ${\textbf{\textit{E}}}_{\rm{S}} \in \mathbb{R}^{\rm{H \times W \times C}}$ generated by the lightweight student autoencoder is used to obtain queries ${\textbf{\textit{Q}}}$ through linear projection, which is defined as:
\begin{equation}
{\textbf{\textit{Q}}} = {\textbf{\textit{W}}}^{\rm{Q}}{\textbf{\textit{E}}}_{\rm{S}} \in \mathbb{R}^{\rm{H \times W \times 2C}}. \label{eq7}
\end{equation}
Here, ${\textbf{\textit{W}}}^{\rm{Q}}$ represents a linear projection learned by a  $1 \times 1$ convolution.

\textbf{Obtaining Keys.}
In order to reduce the computational burden, a $\rm{K} \times \rm{K}$ convolution is applied to the encoded feature map ${\textbf{\textit{E}}}_{\rm{S}}$ prior to obtaining keys ${\textbf{\textit{K}}}$.
\begin{equation}
{\hat{\textbf{\emph{E}}}} = \boldsymbol{\xi}({\textbf{\textit{E}}}_{\rm{S}}) \in \mathbb{R}^{\rm{\hat{H} \times \hat{W} \times C}}. \label{eq8}
\end{equation}
The output of the spatial reduction process is denoted as ${\hat{\textbf{\emph{E}}}}$, where $\boldsymbol{\xi}(\cdot)$ represents the convolution operation, and the stride of the convolution, denoted by $\rm{S}$, is equal to the kernel size $\rm{K}$. The dimensions of the reduced feature map in terms of height and width are represented by $\rm{\hat{H}} = \rm{H}/{S}$ and $\rm{\hat{W}} = \rm{W}/{S}$, respectively. Subsequently, the multiple attention heads are segregated into two groups, to concentrate on spatial and spectral information separately. \Cref{fig:Overview} provides a visual representation of this process.

\emph{Spectral group:}
As each pixel of the hyperspectral image has a unique spectral curve, we split the generated ${\hat{\textbf{\emph{E}}}}$ along the spatial dimension into individual spectral vectors ${\textbf{\textit{S}}} = \{ {\textbf{\textit{S}}}_{i} \in \mathbb{R}^{\rm{1 \times 1 \times C}} | \emph{i}=1,\dots,\rm{\hat{H}\hat{W}} \}$.
Next, we apply linear transformations to each spectral vector and then combine the resulting vectors to obtain the final feature map ${\textbf{\textit{K}}}_{1}$. These keys correspond to spectral information:
\begin{equation}
{\textbf{\textit{K}}}_{1} = \{ {\textbf{\textit{K}}}_{1,\emph{i}} \in \mathbb{R}^{1 \times 1 \times \rm{C}} | \emph{i}=1,\dots,\rm{\hat{H}\hat{W}} \},
{\textbf{\textit{K}}}_{1,\emph{i}} = {\textbf{\textit{W}}}^{\rm{K}_{1}} {\textbf{\textit{S}}}_{i}, \label{eq9}
\end{equation}
where ${\textbf{\textit{W}}}^{\rm{K}_{1}}$ is a parameter matrix learned by a $1 \times 1$ convolution. The above operation involves masking the pixel-to-pixel relationship to concentrate on the spectral information.

\emph{Spatial group:}
To focus on spatial information, ${\hat{\textbf{\emph{E}}}}$ is split along the spectral dimension, generating spatial vectors ${\textbf{\textit{L}}} = \{ {\textbf{\textit{L}}}_{\emph{i}} \in \mathbb{R}^{\rm{\hat{H} \times \hat{W} \times 1}} | \emph{i}=1,\dots,\rm{C} \}$. The keys correspond to spectral information ${\textbf{\textit{W}}}^{\rm{K}_{2}}$ are extracted as follows:
\begin{equation}
{\textbf{\textit{K}}}_{2} = \{ {\textbf{\textit{K}}}_{2,\emph{i}} \in \mathbb{R}^{\rm{\hat{H} \times \hat{W} \times C}} | \emph{i}=1,\dots,\rm{C} \},
{\textbf{\textit{K}}}_{2,\emph{i}} = {\textbf{\textit{W}}}^{\rm{K}_{2}} {\textbf{\textit{L}}}_{i}, \label{eq10}
\end{equation}
where ${\textbf{\textit{W}}}^{\rm{K}_{2}}$ is a parameter matrix learned by $3 \times 3$ convolution. This extraction process blocks any interference caused by the differences between spectral channels.

Following the aforementioned procedures, we obtain the keys as ${\textbf{\textit{K}}} = \{ {\textbf{\textit{K}}}_{\emph{k}} \in \mathbb{R}^{\rm{\hat{H} \times \hat{W} \times C}} | \emph{k}=1,2\}$, where the subscript $k$ denotes the group number of attention heads. Similarly, the values are derived from identical layers but with distinct weights within each group, denoted as ${\textbf{\textit{V}}} = \{ {\textbf{\textit{V}}}_{\emph{k}} \in \mathbb{R}^{\rm{\hat{H} \times \hat{W} \times C}} | \emph{k}=1,2\}$.

\textbf{Spectral-Spatial Multi-head Attention.}
The MSS block utilizes a novel spectral-spatial multi-head attention mechanism, which involves splitting the queries ${\textbf{\textit{Q}}}$ along the channel dimension into two distinct parts, denoted as  ${\textbf{\textit{Q}}} = \{ {\textbf{\textit{Q}}}_{\emph{k}} \in \mathbb{R}^{\rm{H \times W \times C}} | \emph{k}=1,2\}$ and assigning them to separate head groups. Next, each head group performs self-attention to learn either spectral or spatial information. The attended feature ${\textbf{\textit{A}}}_{\emph{k}}$ for each group is calculated as follows:
\begin{equation}
{\textbf{\textit{A}}}_{\emph{k}} = {\bf\emph{Softmax}}(\frac{{\textbf{\textit{Q}}}_{\emph{k}}{\textbf{\textit{K}}}_{\emph{k}}^{\top}}{\sqrt{\emph{h}}}){\textbf{\textit{V}}}_{\emph{k}} \in \mathbb{R}^{\rm{H \times W \times C}},{\emph{k}=1,2}, \label{eq11}
\end{equation}
where $\sqrt{\emph{h}}$ is a scaling factor. Then, aggregating the attended features from the two head groups allows for information interaction:
\begin{equation}
{\textbf{\textit{A}}} = \boldsymbol{\varphi}(\boldsymbol{\delta}({\textbf{\textit{A}}}_{1},{\textbf{\textit{A}}}_{2})) \in \mathbb{R}^{\rm{H \times W \times C}}, \label{eq12}
\end{equation}
$\boldsymbol{\delta}(\cdot)$ represents the concatenation operation and $\boldsymbol{\varphi}(\cdot)$ represents
a linear layer along the channel dimension. The output feature map ${\textbf{\textit{F}}} \in \mathbb{R}^{\rm{{H} \times {W} \times C}}$ of the MSS block can be calculated as:
\begin{equation}
{\textbf{\textit{F}}} = \boldsymbol{\eta}({\textbf{\textit{A}}}) + {\textbf{\textit{E}}}_{\rm{S}}, \label{eq13}
\end{equation}
where $\boldsymbol{\eta}(\cdot)$ represents a feed-forward network implemented by multi-layer perceptron.

\subsubsection{Saliency Map Generation}
The MSST integrated the hierarchical feature maps generated by multiple stages into an FPN structure to produce high-quality saliency maps, which effectively combined multi-scale features. The final saliency map, denoted as $\textbf{\textit{Y}}$, was computed as follows:
\begin{align}
&{\textbf{\textit{Y}}} = {\boldsymbol{\omega}}(\boldsymbol{\sigma}({\textbf{\textit{F}}}_{1},{\textbf{\textit{F}}}'_{2})), \\
&{\textbf{\textit{F}}}'_{2} = \boldsymbol{\sigma}({\textbf{\textit{F}}}_{2},{\textbf{\textit{F}}}'_{3}), \\
&{\textbf{\textit{F}}}'_{3} = \boldsymbol{\sigma}({\textbf{\textit{F}}}_{3},{\textbf{\textit{F}}}_{4}). \label{eq14}
\end{align}
Here, ${\textbf{\textit{F}}}_{\emph{i}}$ represents the feature map from the $i$-th stage, and ${\textbf{\textit{F}}}'_{\emph{i}}$ represents the feature map generated by the $i$-th layer in the FPN structure. $\boldsymbol{\omega}(\cdot)$ represents the convolution operation. The fusion operation $\boldsymbol{\sigma}(\cdot,\cdot)$ of feature maps is expressed as:
\begin{equation}
\bm{\sigma}({\textbf{\textit{X}}},{\textbf{\textit{Y}}}) = {\boldsymbol{\omega}}({\textbf{\textit{X}}}) + \boldsymbol{\theta}({\textbf{\textit{Y}}}), \label{eq15}
\end{equation}
where $\boldsymbol{\theta}(\cdot)$ represents an upsampling operation with a magnification factor of $2$.

\subsection{Learning Objectives}
Our proposed DMSSN is trained in a joint end-to-end manner, and the network is optimized using a hybrid loss function that comprises three components: a data reconstruction loss $\mathcal{L}_{\rm{S}}$, a saliency detection loss $\mathcal{L}_{\rm{sod}}$, and a distillation loss $\mathcal{L}_{\rm{dis}}$. The hybrid loss function is expressed as:
\begin{equation}
\mathcal{L} = \mathcal{L}_{\rm{S}} + \mathcal{L}_{\rm{sod}} + \mathcal{L}_{\rm{dis}}. \label{eq16}
\end{equation}
The following section provides a detailed explanation of these loss functions and the training process.

\subsubsection{Pre-training Loss}
The deep teacher autoencoder undergoes pre-training using the loss function $\mathcal{L}_{\rm{T}}$, which is defined as follows:
\begin{equation}
\mathcal{L}_{\rm{T}} = \mathcal{L}_{\rm{hs}}({\textbf{\textit{D}}}_{\rm{T}},{\textbf{\textit{I}}}), \label{eq17}
\end{equation}
where $\mathcal{L}_{\rm{hs}}(\cdot,\cdot)$ is a combination of Huber and Spectral Angle Mapping (SAM) computed between the decoding result of the teacher autoencoder ${\textbf{\textit{D}}}_{\rm{T}}$ and the input hyperspectral image ${\textbf{\textit{I}}}$. The pre-training phase using $\mathcal{L}_{\rm{T}}$ allows the deep teacher autoencoder to learn a powerful encoding capability and gain valuable knowledge about the hyperspectral data. This knowledge will later be transferred to the lightweight student autoencoder during the knowledge distillation process, contributing to efficient dimensionality reduction while preserving important information.

\subsubsection{Distillation Loss}
To enhance the encoding capability of the student autoencoder, we compute the loss function $\mathcal{L}_{\rm{S}}$ between the data reconstruction results generated by the autoencoder in the decoding stage ${\textbf{\textit{D}}}_{\rm{S}}$ and the input hyperspectral image ${\textbf{\textit{I}}}$, as the decoding results are only associated with the encoding results:
\begin{equation}
\mathcal{L}_{\rm{S}} = \mathcal{L}_{\rm{hs}}({\textbf{\textit{D}}}_{\rm{S}},{\textbf{\textit{I}}}). \label{eq18}
\end{equation}

With the help of the knowledge distillation strategy, the lightweight student autoencoder possesses a similar encoding ability as the deep teacher autoencoder without incurring additional computational costs. The teacher autoencoder loads the pre-trained model and freezes the parameters to generate feature maps corresponding to three guidance layers. Then, the distillation loss between three groups of feature maps can be calculated as follows:
\begin{equation}
\begin{aligned}
\mathcal{L}_{\rm{dis}} &= \mathcal{L}_{\rm{hs}}({\textbf{\textit{E}}}_{\rm{T}}^{2},{\textbf{\textit{E}}}_{\rm{S}}^{1}) + \mathcal{L}_{\rm{hs}}({\textbf{\textit{E}}}_{\rm{T}},{\textbf{\textit{E}}}_{\rm{S}})\\
& + \mathcal{L}_{\rm{hs}}({\textbf{\textit{D}}}_{\rm{T}}^{2},{\textbf{\textit{D}}}_{\rm{S}}^{1}), \label{eq19}
\end{aligned}
\end{equation}
where ${\rm{T}}$ and ${\rm{S}}$ represent the teacher and student autoencoders. The first set is the encoding intermediate results, the second set is the final encoding results, and the last set is the decoding intermediate results. 

The distillation loss $\mathcal{L}_{\rm{dis}}$ encourages the student autoencoder to mimic the encoding and decoding process of the teacher autoencoder and learn similar feature representations. By minimizing this loss, the student autoencoder effectively inherits the knowledge acquired by the deep teacher autoencoder, leading to improved encoding capabilities and higher performance in the hyperspectral salient object detection task.

\subsubsection{Saliency Loss}
During the training process, the DMSSN employs a constraint to ensure model convergence by computing the deviation between the final saliency map ${\textbf{\textit{Y}}}$ and the corresponding ground truth ${\textbf{\textit{T}}}$. To achieve this, a loss function $\mathcal{L}_{\rm{sod}}$ is utilized, which is comprised of three components: Binary Cross-Entropy Loss (BCE), Structural Similarity Index (SSIM), and Intersection over Union (IOU). The loss function is defined as follows:
\begin{equation}
\begin{aligned}
\mathcal{L}_{\rm{sod}} &= \mathcal{L}_{\rm{bce}}({\textbf{\textit{Y}}},{\textbf{\textit{T}}}) + \mathcal{L}_{\rm{ssim}}({\textbf{\textit{Y}}},{\textbf{\textit{T}}}) \\
&+ \mathcal{L}_{\rm{iou}}({\textbf{\textit{Y}}},{\textbf{\textit{T}}}). \label{eq20}
\end{aligned}
\end{equation}
Here, $\mathcal{L}_{\rm{bce}}(\cdot,\cdot)$, $\mathcal{L}_{\rm{ssim}}(\cdot,\cdot)$, and $\mathcal{L}_{\rm{iou}}(\cdot,\cdot)$  correspond to BCE, SSIM, and IOU loss functions, respectively. By minimizing the loss function $\mathcal{L}_{\rm{sod}}$, the DMSSN learns to generate accurate and visually consistent saliency maps that closely resemble the ground truth, leading to improved performance in the hyperspectral salient object detection task.

By combining these three loss components, the hybrid loss function enables comprehensive and effective training of the DMSSN. The joint end-to-end training approach ensures that all components of the network work harmoniously, leading to superior performance in hyperspectral salient object detection tasks.

\begin{figure*}[t]
  \centering
  \includegraphics[width=1\linewidth]{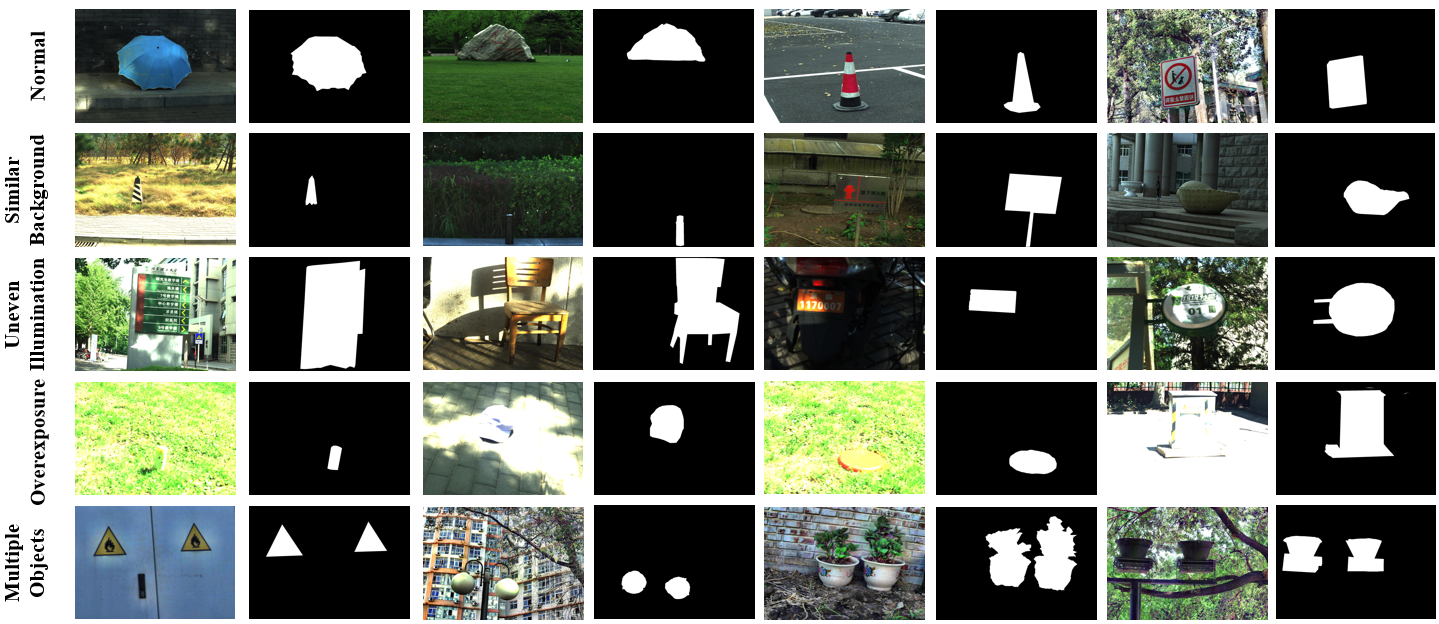}
   \caption{Examples of pseudo-color images and ground truth annotations from the HSOD-BIT dataset. Some of the collected scenes in this dataset cannot be adequately processed by RGB methods, which highlights the need for hyperspectral data in the HSOD task.}
   \label{fig:HSOD-C}
\end{figure*}

\section{HSOD-BIT Dataset}


\noindent\textbf{Collection Platform.}
We employed the Norsk Elektro Optikk A/S HySpex Mjolnir-1024 hyperspectral imaging camera to acquire our data, and utilized additional equipment, including the DEREE C300 platform, DEREE Ts-16 tripod, and STANDA 8MR190 precision electric turntable. The hyperspectral camera and turntable are used in tandem to conduct line scan imaging of natural scenes. The dataset includes data collected from various locations on the Beijing Institute of Technology campus. We conducted data collection across multiple weather conditions and periods to ensure the diversity and representativeness of the scenes in the dataset. Throughout the data collection process, we kept imaging parameters such as exposure time, white balance, and gain compensation constant to maintain consistency in the data. 

\noindent\textbf{Data Annotation.}
The HSOD-BIT dataset comprises a total of $366$ hyperspectral images. To ensure data quality, the images underwent a thorough evaluation by multiple trained observers. Images with issues such as motion blur or inappropriate object sizes were excluded from the dataset, resulting in a final selection of $319$ high-quality hyperspectral images for the HSOD task.
The diversity of object types and sizes is ensured in both the training and test sets, with the former consisting of $255$ images and the latter containing $64$ images. Pixel-level ground truth labels were manually created for each image using the ImageLabeler toolkit. Trained annotators carefully labeled each pixel as either belonging to the target object or the background. \Cref{fig:HSOD-C} displays some example hyperspectral images from the HSOD-BIT dataset along with their corresponding pseudo-color representations and ground truth annotations. 

\noindent\textbf{Complex Conditions Challenge.}
To highlight the advantages of hyperspectral images, the HSOD dataset includes images captured under complex conditions, such as a similar background, uneven illumination, and overexposure, which are difficult for traditional RGB images composed of three primary colors to handle. As shown in \Cref{fig:HSOD-C}, these challenges make it difficult to distinguish the target boundary or even determine the object's position using traditional RGB-based SOD. In contrast, hyperspectral images provide specific spectral information for each pixel, allowing HSOD methods to effectively capture the spectral signatures of the target objects and the background. As a result, HSOD is more promising and effective compared to traditional RGB-based SOD, especially when dealing with intricate scenarios with similar backgrounds, uneven lighting, and overexposure. The HSOD-BIT dataset serves as an ideal platform for evaluating and advancing the performance of HSOD methods in such challenging conditions.

\begin{table}[t]
\caption{Statistics of Our HSOD-BIT and Previous HS-SOD Dataset}
\centering
\label{tab:Comparison}
\begin{tabular}{l|cc}
    \toprule
    Property & HS-SOD \cite{imamoglu2018hyperspectral} & HSOD-BIT \\
    \midrule
    Data Volume & $60$ & $319$ \\
    Spatial Resolution & $768 \times 1024$ & $1240 \times 1680$ \\
    Spectral Bands & $81$ & $200$ \\
    Spectral Resolution & $5$ nm & $3$ nm \\
    Spectral Range & $380$ -- $700$ nm & $400$ -- $1000$ nm \\
    \bottomrule
  \end{tabular}
\end{table}

\noindent\textbf{Data Statistics.}
\Cref{tab:Comparison} presents a comparison between the proposed HSOD-BIT and the commonly used HS-SOD \cite{imamoglu2018hyperspectral}.
HSOD-BIT, which consists of $319$ hyperspectral images, is more prominent in scale and has an order of magnitude more data than HS-SOD. In terms of spatial resolution, all images in HSOD-BIT are cropped to a high resolution of $1240 \times 1680$, providing more detailed spatial information compared to HS-SOD. The spectral resolution of HSOD-BIT is also significantly higher than HS-SOD. HSOD-BIT stores all $200$ spectral bands captured by the hyperspectral camera, ranging from visible to near-infrared, offering a more comprehensive and detailed representation of the spectral information of the scenes. Overall, HSOD-BIT outperforms HS-SOD in terms of dataset size, spatial resolution, and spectral resolution, making it a valuable resource for advancing hyperspectral salient object detection research.

\begin{figure}[t]
  \centering
   \includegraphics[width=1\linewidth]{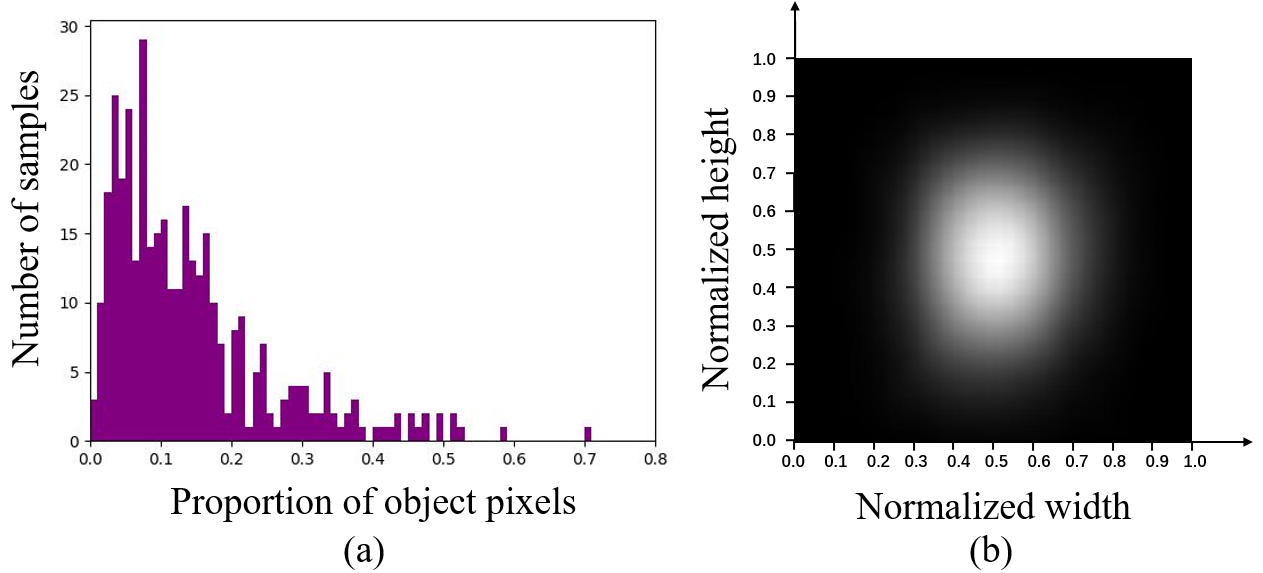}
   \caption{Diagram of HSOD-BIT statistics. (a) The distribution of the number of images according to different object scales. (b) The probability density function of the spatial distribution of object centroids.}
   \label{fig:statistic}
\end{figure}

The analysis of object scales in the HSOD-BIT dataset provides valuable insights into the diversity of object sizes present in the images. As depicted in \Cref{fig:statistic}(a), the majority of images contain small and medium-scale objects, with a relatively even distribution across these scales. Additionally, a significant proportion of images ($22.4\%$) feature large-scale objects, defined as objects with more than $30\%$ of the image area. This diversification of object scales in the dataset is beneficial for training robust HSOD methods. By including images with objects of varying sizes, the method becomes more versatile and capable of accurately detecting salient objects across different scales. This is particularly important in real-world scenarios where objects can appear in various sizes and at different distances from the camera.

We perform an object centroid analysis to count the spatial distribution of objects in each image and visualize the probability distribution of their centroids in the dataset. 
As shown in \Cref{fig:statistic}(b), the high brightness regions in the probability distribution indicate a dense concentration of object centroids. \Cref{fig:statistic}(b) illustrates the probability distribution of object centroids, where regions with higher brightness indicate a denser distribution of centroids. It is evident from the visualization that the centroids tend to be concentrated in the center of the images. This finding aligns well with the natural tendency of the human eye to focus on salient objects in the central region of the field of view. This characteristic is essential to consider when developing HSOD methods, as it reflects the way humans perceive and attend to visual stimuli. By analyzing the spatial distribution of object centroids, we gain a deeper understanding of the dataset's characteristics, which can guide the development of more effective and accurate HSOD methods.

\begin{figure}[t]
  \centering
   \includegraphics[width=1\linewidth]{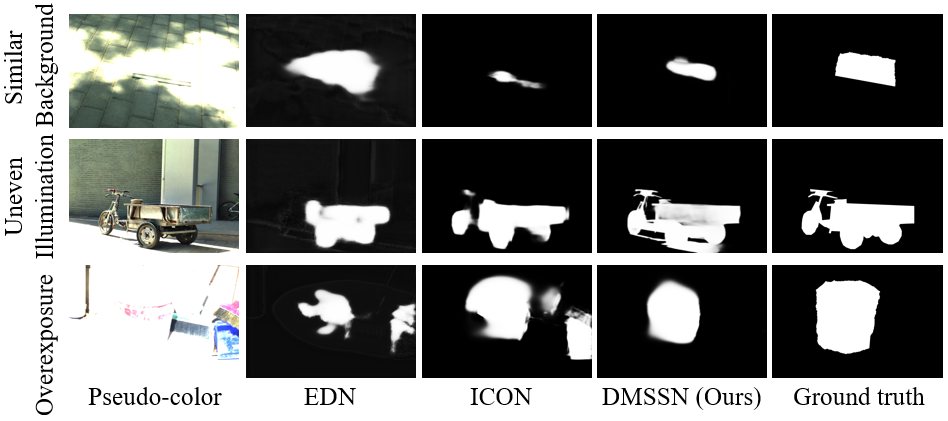}
   \caption{Comparison of prediction results between RGB-based and HSI-based methods on the HSOD-BIT dataset under complex conditions.}
   \label{fig:Comparison}
\end{figure}

\section{Experiment}

\subsection{Implementation Details}\label{subsec5.1}
The Distilled Mixed Spectral-Spatial Network (DMSSN) is trained from scratch on the HSOD-BIT dataset. During the pre-training stage of the teacher autoencoder, an AdamW optimizer is adopted with an initial learning rate of $0.002$, a batch size of $4$, and $50$ epochs. The complete DMSSN is trained for $100$ epochs under the guidance of the teacher autoencoder using an AdamW optimizer with an initial learning rate of $0.06$ and a batch size of $12$. Random selection, random scaling, and center cropping are used as data augmentation techniques to expand the training samples. 

To comprehensively evaluate the performance of DMSSN, various popular SOD evaluation metrics \cite{li2023delving, wang2017salient}, including Mean Absolute Error (MAE), Precision (PRE), Recall (REC), average $\rm{F}_1$-Measure (avg$\rm{F}_1$), Area Under Curve (AUC), Correlation Coefficient (CC), and Normalized Scanpath Saliency (NSS). By reporting these evaluation metrics, the performance of DMSSN can be comprehensively assessed and compared with other HSOD methods to gauge its effectiveness in salient object detection in hyperspectral images.

\begin{table*}[t]
\caption{Quantitative Comparison with State-of-the-art RGB-based and HSI-based Methods on HSOD-BIT}
\centering
\label{tab:ReHLSOD}
  \setlength{\tabcolsep}{8pt}
  \begin{tabular}{c|l|ccccccc|cc}
    \toprule
    & Method & MAE $\downarrow$ & PRE $\uparrow$ & REC $\uparrow$ & avg$\rm{F}_1$ $\uparrow$ & AUC $\uparrow$ & CC $\uparrow$ & NSS $\uparrow$ & \#Params(M) & FLOPs(G)\\
    \midrule
    \multirow{6}{*}{RGB-based} & Itti \cite{itti1998model} & 0.274 & 0.181 & 0.239 & 0.152 & 0.565 & 0.131 & 0.589 & - & -\\
    & EDN \cite{wu2022edn} & 0.120 & 0.817& 0.361 & 0.463 & 0.664 & 0.462 & 1.461 & 33.04 & 20.42\\
    & BASNet \cite{qin2019basnet} & 0.090 & 0.544 & 0.577 & 0.477 & 0.761 & 0.479 & 1.547 & 87.06 & 127.56 \\
    & U2Net \cite{qin2020u2} & 0.072 & 0.637 & 0.656 & 0.589 & 0.804 & 0.613 & 1.998 & 44.01 & 47.65 \\
    & ICON \cite{zhuge2022salient} & 0.071 & 0.479 & 0.830 & 0.588 & 0.818 & 0.606 & 2.107 & 33.09 & 8.49 \\
    & PSOD \cite{gao2022weakly} & 0.086 & 0.796 & 0.553 & 0.632 & 0.760 & 0.621 & 2.069 & 91.61 & 108.89\\
    \midrule
    \multirow{7}{*}{HSI-based} & GS \cite{yan2016salient} & 0.161 & 0.159 & 0.724 & 0.228 & 0.795 & 0.509 & 1.647 & - & -\\
    & SAD \cite{yan2016salient} & 0.204 & 0.182 & 0.469 & 0.224 & 0.668 & 0.328 & 1.065 & - & -\\
    & SED \cite{yan2016salient} & 0.146 & 0.046 & 0.686 & 0.076 & 0.772 & 0.331 & 1.084 & - & -\\
    & SED-GS \cite{yan2016salient} & 0.186 & 0.134 & 0.712 & 0.196 & 0.789 & 0.434 & 1.424 & - & -\\
    & SED-SAD \cite{yan2016salient} & 0.216 & 0.147 & 0.672 & 0.207 & 0.770 & 0.384 & 1.270 & - & -\\
    & SUDF \cite{imamouglu2019salient} & 0.149 & 0.525 & 0.637 & 0.485 & 0.790 & 0.625 & 1.920 & 0.10 & 82.90\\ 
    \cmidrule{2-11}
    & DMSSN (Ours) & 0.086 & 0.663 & 0.676 & 0.621 & 0.817 & 0.636 & 2.171 & 1.76 & 10.89\\
    \bottomrule
  \end{tabular}
\end{table*}

\subsection{Results On HSOD-BIT}

\subsubsection{Quantitative Analysis}
\Cref{tab:ReHLSOD} presents a comparative analysis of various methods for HSOD tasks. The findings highlight the superior performance of deep methods over traditional methods, particularly in metrics such as CC and NSS. Notably, the proposed DMSSN exhibits superior detection performance compared to the deep method SUDF. For instance, DMSSN improves AUC by $0.027$, CC by $0.011$, and NSS by $0.251$. These improvements demonstrate the efficacy of DMSSN in accurately detecting salient objects. 

Notably, the REC of DMSSN is not exceptional, even lower than that of some traditional methods. But at the same time we noticed that, except DMSSN, the PRE of other methods is much lower than REC. This situation essentially reflects that these methods have misjudged the background as the target.
When considering both PRE and REC, avg$\rm{F}_1$ provides a balanced assessment, which demonstrates that DMSSN reaches a maximum of $0.621$.
The quantitative analysis confirms that the proposed DMSSN achieves state-of-the-art performance for HSOD, outperforming existing deep methods and traditional methods in various evaluation metrics.

\begin{figure}[t]
  \centering
   \includegraphics[width=1\linewidth]{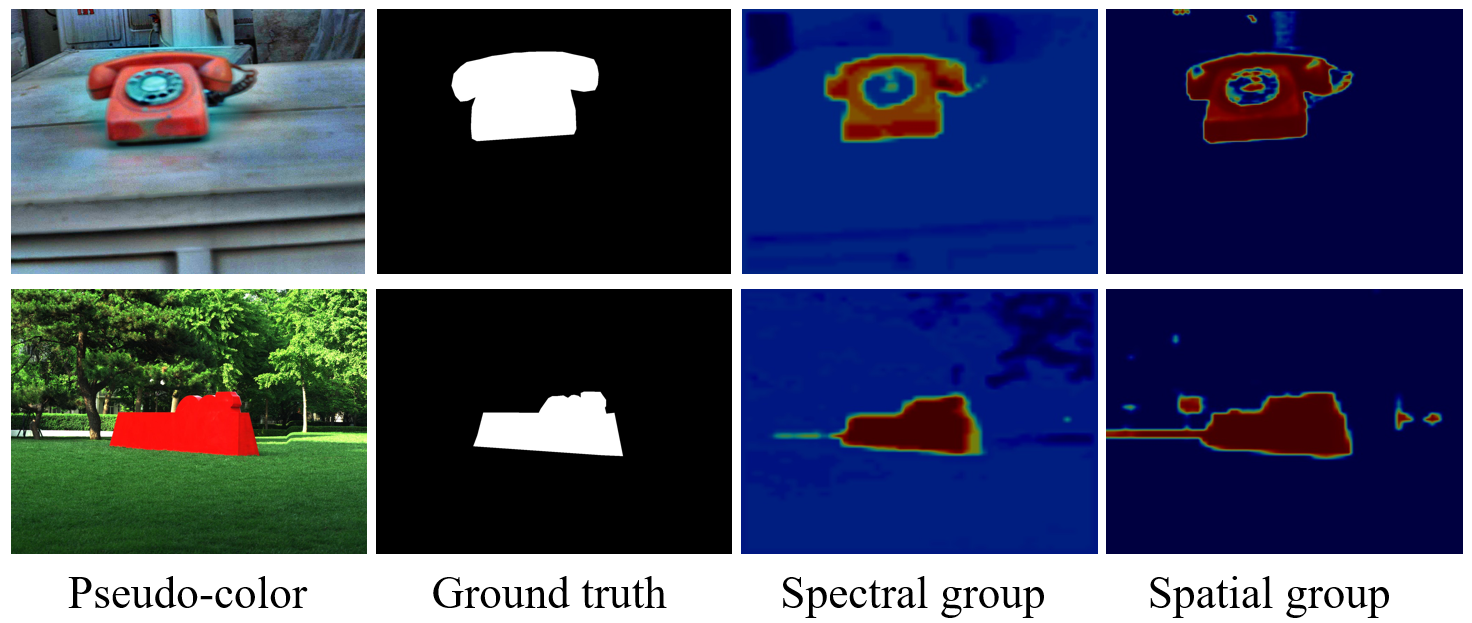}
   \caption{Visualization of the feature maps extracted separately by the two attention head groups in the MSS block.}
   \label{fig:spespa}
\end{figure}

\subsubsection{Comparison with RGB-based Methods}
We evaluate the performance of the proposed HSI-based DMSSN against several state-of-the-art RGB-based SOD methods \cite{itti1998model, wu2022edn, qin2019basnet, qin2020u2, zhuge2022salient, gao2022weakly} under complex conditions. \Cref{fig:Comparison} illustrates that the detection results of RGB-based methods exhibit a notable bias, while the performance of HSI-based methods, particularly DMSSN, remains stable and robust. In scenarios with uneven illumination, RGB-based methods tend to produce blurred edges and lose track of the target. Conversely, the proposed HSI-based DMSSN accurately delineates the object contour, benefiting from the full extraction of spectral-spatial features. Under challenging conditions like overexposure and similar background scenes, where RGB-based methods struggle to determine the object position and may lose track entirely, DMSSN consistently achieves high-quality salient object detection, showcasing its robust representation ability based on hyperspectral images.

The quantitative analysis presented in \Cref{tab:ReHLSOD} further confirms the poor detection performance of RGB-based methods on the HSOD-BIT dataset. These methods significantly lag behind the proposed DMSSN in all metrics and even perform worse than traditional HSI-based methods on certain metrics. In conclusion, the HSI-based DMSSN demonstrates greater potential and superiority compared to RGB-based methods in the HSOD task, especially under various challenging complex conditions such as uneven illumination, overexposure, and similar background scenes.

\begin{figure*}[t]
  \centering
   \includegraphics[width=1\linewidth]{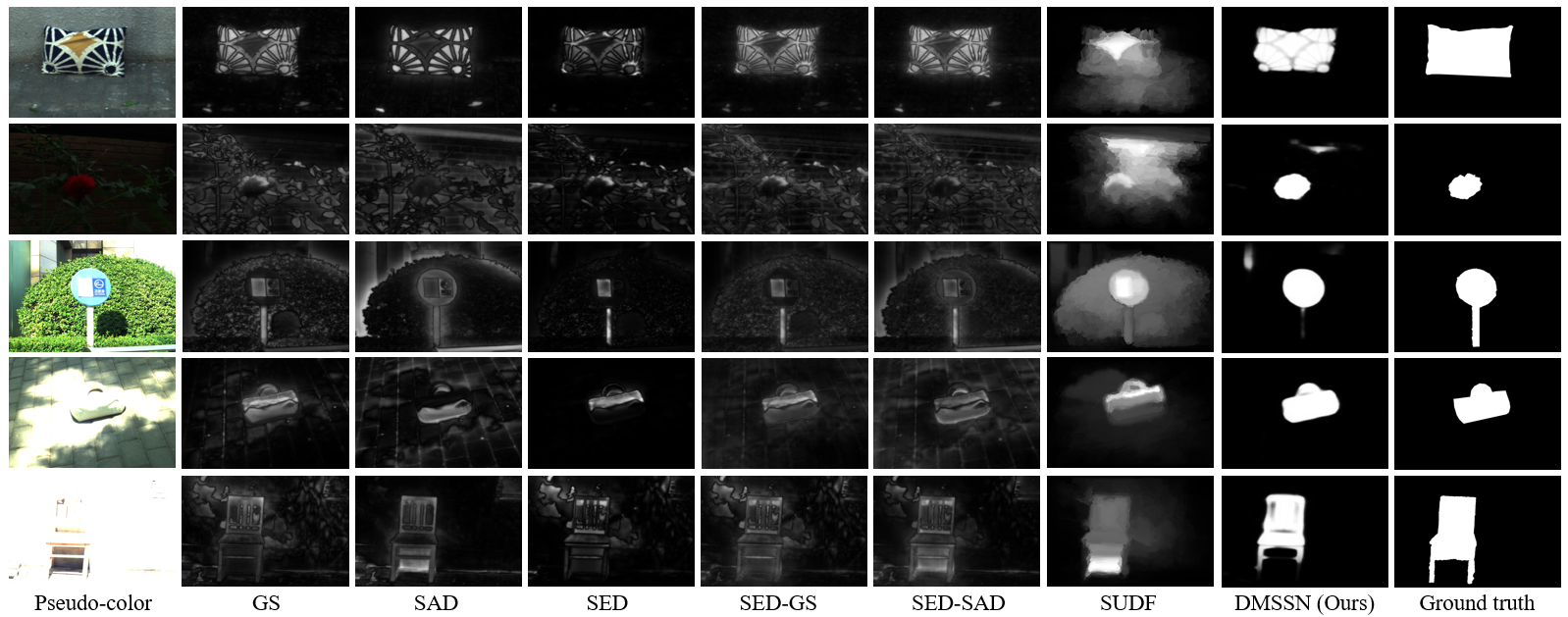}
   \caption{Detection results of multiple HSI-based methods on the HSOD-BIT dataset.}
   \label{fig:HLresult}
\end{figure*}

\subsubsection{Comparison of efficiency}
We conducted a comparative analysis of the efficiency of the proposed DMSSN and other deep-learning methods, including model parameters (\#Param) and floating point operations (FLOPs). \Cref{tab:ReHLSOD} shows the results of each method under the default configuration, and the traditional method is blank. Obviously, the RGB-based methods do not consider the spectral dimension and usually adopt complex network structures, resulting in high model parameters. In contrast, HSI-based methods have fewer parameters to accommodate hyperspectral data with high spectral resolution. Notably, the SUDF only employs a CNN for feature extraction, followed by manifold learning and superpixel clustering. Consequently, it has minimal parameters but high FLOPs. Compared with previous methods, the proposed DMSSN adopts the knowledge distillation strategy, achieving strong modeling capability with low parameters and FLOPs. The above comparisons demonstrate that DMSSN offers a trade-off between computational efficiency and performance.

\subsubsection{Visualization of Attention Maps}
\Cref{fig:spespa} illustrates the feature maps of each attention group in the proposed MSS block, which incorporates both spectral and spatial information through a multi-head attention mechanism.
The spectral group focuses on capturing the spectral characteristics of the hyperspectral image, highlighting the target based on the significant spectral contrast between the salient object and the background. 
The spatial group, on the other hand, emphasizes the spatial relationships between pixels and contributes to better delineating the boundaries of the salient regions.
By aggregating the feature maps from the two groups, the MSS block effectively combines both spectral and spatial information. This fusion of complementary features results in a saliency map with high precision, accurately locating the salient objects and providing a more detailed representation of their boundaries.

\subsubsection{Visualization of Predicted Saliency Maps}
\Cref{fig:HLresult} provides a visual comparison of pseudo-color images, ground truths, and saliency maps from multiple HSI-based methods. The comparison reveals the varying detection capabilities of these methods under different conditions. Conventional methods, as shown in the figure, have limited detection capabilities and can only offer a rough outline of salient objects. This deficiency severely impacts the overall detection accuracy, and these methods are prone to misidentifying the background as the salient object, especially under complex and challenging conditions. In contrast, deep methods significantly improve detection performance by accurately highlighting the positions of salient objects. Compared to the SUDF method, the proposed DMSSN leverages both spectral and spatial information concurrently, resulting in a remarkable ability to differentiate between the background and salient objects. The saliency maps produced by DMSSN exhibit sharper contours and more accurate representations of the target objects.

\Cref{fig:HLresult} further demonstrates that the proposed DMSSN achieves exceptional prediction results under diverse conditions, including conventional scenes, similar backgrounds, overexposure, and uneven illumination. This robustness and superior performance make DMSSN a promising solution for hyperspectral salient object detection across various environmental conditions and challenges.

\begin{figure*}[t]
  \centering
   \includegraphics[width=1\linewidth]{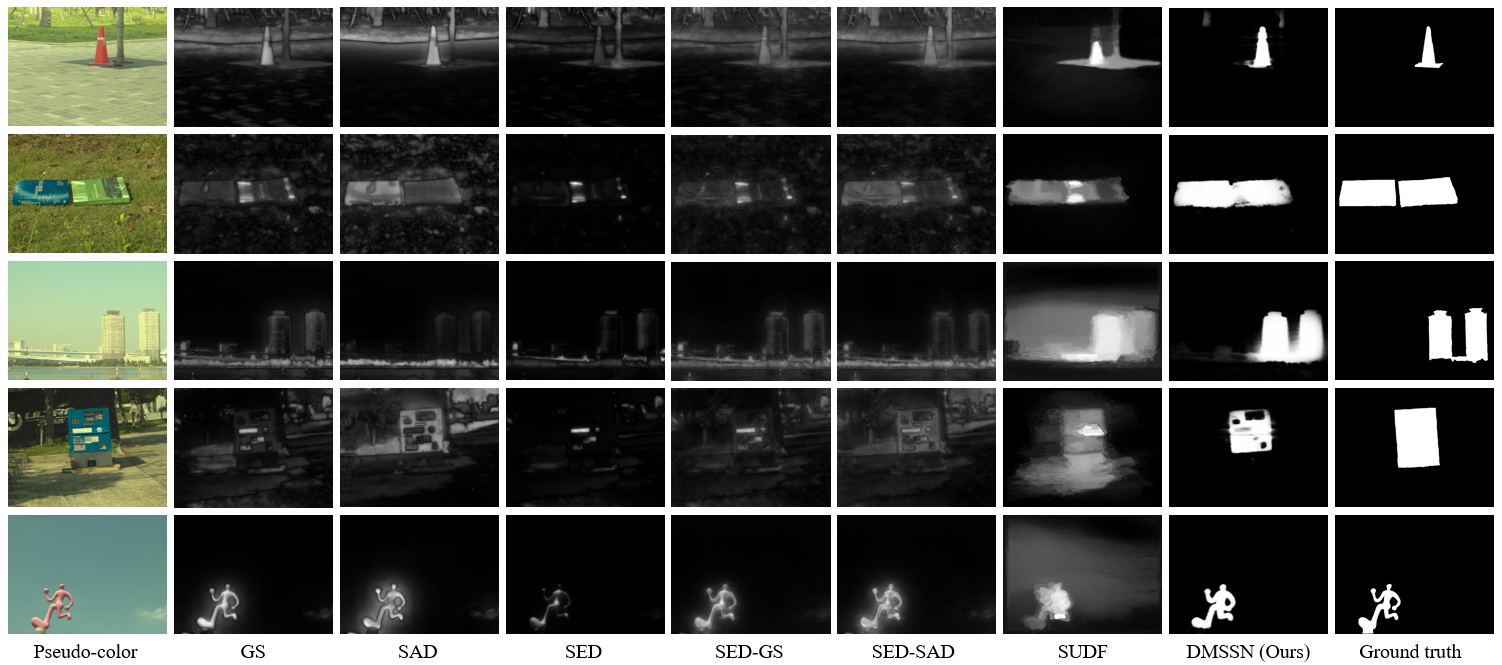}
   \caption{Detection results of multiple HSI-based methods on HS-SOD.}
   \label{fig:HSresult}
\end{figure*}

\subsection{Results on HS-SOD}

\subsubsection{Quantitative Analysis}
\Cref{tab:ReHSSOD} presents a comparison of the proposed DMSSN with other commonly used HSI-based methods on the HS-SOD \cite{imamoglu2018hyperspectral} dataset. To ensure a fair comparison, we maintain the training settings consistent with previous work \cite{huang2021salient}, which employs $48$ hyperspectral images for training and the remaining samples for testing. We fine-tune a pre-trained model from HSOD-BIT with an initial learning rate of $0.002$, a batch size of $12$, and a total of $50$ epochs. The results demonstrate that the proposed DMSSN outperforms other popular HSI-based methods and achieves state-of-the-art performance, with a slightly smaller AUC compared to HSISO but improved performance in terms of MAE and avg$\rm{F}_1$.

\subsubsection{Qualitative Analysis}
\Cref{fig:HSresult} provides a visualization of the prediction results for the compared HSI-based methods on the HS-SOD dataset. The visualization confirms that the proposed DMSSN exhibits the most reliable and accurate detection performance compared to other HSI-based methods.
DMSSN demonstrates a robust ability to suppress background interference and accurately highlight salient objects, resulting in saliency maps with sharp and precise contours. It effectively captures and utilizes the spectral and spatial information present in hyperspectral images, enabling it to better distinguish between salient objects and the background.
In contrast, other HSI-based methods show limitations in their ability to accurately detect salient objects. Some methods produce saliency maps with blurred edges, indicating a weaker ability to differentiate between salient objects and the background. Others may misjudge the background as the object, leading to less accurate and reliable predictions.
Overall, the visualization results reinforce the superior performance of DMSSN in hyperspectral salient object detection.

\begin{table}[t]
\caption{Results on Previous HS-SOD Dataset.}
\centering
  \begin{tabular}{l|ccccc}
    \toprule
    Method & MAE $\downarrow$ & avg$\rm{F}_1$ $\uparrow$ & AUC $\uparrow$ & CC $\uparrow$ & NSS $\uparrow$ \\
    \midrule
    GS \cite{yan2016salient} & 0.217 & 0.290 & 0.778 & 0.375 & 1.607 \\
    SAD \cite{yan2016salient} & 0.235 & 0.269 & 0.771 & 0.377 & 1.548 \\
    SED \cite{yan2016salient} & 0.182 & 0.168 & 0.769 & 0.369 & 1.549 \\
    SED-GS \cite{yan2016salient} & 0.183 & 0.188 & 0.802 & 0.382 & 1.688 \\
    SED-SAD \cite{yan2016salient} & 0.206 & 0.168 & 0.811 & 0.384 & 1.627 \\
    SUDF \cite{imamouglu2019salient} & 0.135 & 0.467 & 0.860 & 0.583 & 2.120 \\
    HSISO \cite{huang2021salient} & 0.121 & 0.527 & \textbf{0.907} & 0.555 & 2.038 \\
    \midrule
    DMSSN (Ours) & \textbf{0.068} & \textbf{0.617} & 0.903 & \textbf{0.665} & \textbf{2.040} \\
    \bottomrule
  \end{tabular}
  \label{tab:ReHSSOD}
\end{table}

\subsection{Results Analysis}
The above comprehensive analysis demonstrates that the proposed DMSSN has achieved state-of-the-art performance on multiple HSOD datasets. The outstanding results obtained by DMSSN can be attributed to several innovative features that set it apart from other methods. 

Firstly, the Distilled Spectral Encoding process plays a pivotal role in enhancing the robustness of DMSSN. By reducing background complexity and eliminating the influence of background clutter, this process allows the method to focus on salient objects and extract accurate contours. Additionally, the spectral encoding process efficiently preserves crucial spectral information, which is vital for hyperspectral image analysis. Secondly, the adoption of MSS blocks ensures the effective extraction and full utilization of spectral-spatial features in DMSSN. By simultaneously considering both spectral and spatial information, the method can generate saliency maps with high precision and accuracy. This feature enables DMSSN to outperform other methods in terms of various evaluation metrics. 

The excellent performance of DMSSN is evident in its AUC of $0.817$ and an avg$\rm{F}_1$ of $0.621$ on HSOD-BIT, surpassing other methods. Similarly, on HS-SOD, DMSSN achieves the highest AUC of $0.903$ and an avg$\rm{F}_1$ of $0.617$. The visualization results on both datasets further validate the strong robustness and accurate focus on salient objects achieved by DMSSN. In summary, the proposed DMSSN has demonstrated its superiority in both HSOD-BIT and HS-SOD datasets. Its innovative features and effective utilization of spectral and spatial information have contributed to its outstanding performance and made it advance in this challenging task.

\begin{table}[t]
\caption{Remote sensing target detection results of the proposed DMSSN and other state-of-the-art methods on the San Diego dataset}
\centering
\setlength{\tabcolsep}{10pt}
  \begin{tabular}{l|ccc}
    \toprule
     Method &  $\rm{AUC_{(\tau,P_{d})}}$ & $\rm{AUC_{OA}}$ & Time(s) \\
    \midrule
    SAM \cite{kruse1993spectral}  & 0.9730 & 1.3508 & 0.01 \\
    CEM \cite{farrand1997mapping} & 0.5277 & 1.2249 & 0.02 \\
    hCEM \cite{zou2015hierarchical}  & 0.7421 & 1.4930 & 0.17 \\
    E-CEM \cite{zhao2019ensemble}  & 0.5324 & 1.5200 & 9.23\\
    CSCR \cite{li2015combined} & 0.7689 & 1.6039 & 20.40\\
    DM-BDL \cite{cheng2021decomposition}  & 0.5335 & 1.4784 & 5.07 \\
    HTD-Net \cite{zhang2020htd}  & 0.1773  & 1.1200 & 30.87 \\
    BLTSC \cite{xie2020background} & 0.5326 & 1.5125 & 1.37 \\
    HTD-IRN \cite{shen2023hyperspectral} & 0.5328  & 1.5279 & 0.26 \\
    \midrule
    DMSSN-p(Ours) & 0.4296 & 1.4720 & 1.07 \\
    DMSSN-s(Ours)& 0.6203 & 1.4855 & 1.20 \\
    DMSSN(Ours) & 0.6293 & 1.5365 & 0.79 \\
    \bottomrule
  \end{tabular}
  \label{tab:RSTD}
\end{table}

\begin{table*}[t]
\caption{Applying Mixed Spectral-Spatial Transformer to the Hyperspectral Remote Sensing Images Classification Task}
\centering
\setlength{\tabcolsep}{10pt}
  \begin{tabular}{l|cc|cc|cc|cc}
    \toprule
     \multirow{2}{*}{Method} & \multicolumn{2}{c|}{IP \cite{hong2021spectralformer}} & \multicolumn{2}{c|}{PU \cite{zou2022lessformer}} & \multicolumn{2}{c|}{KSC \cite{yan2021global}} & \multicolumn{2}{c}{WHU-OHS \cite{li2022whu}}\\
     & OA $\uparrow$ & Kappa $\uparrow$ & OA $\uparrow$ & Kappa $\uparrow$ & OA $\uparrow$ & Kappa $\uparrow$ & OA $\uparrow$ & Kappa $\uparrow$ \\
    \midrule
    1D-CNN \cite{hu2015deep} & 0.902 & 0.900 & 0.912 & 0.883 & 0.868 & 0.853 & 0.636 & 0.526 \\
    3D-CNN \cite{li2017spectral} & 0.968 & 0.964 & 0.964 & 0.950 & 0.973 & 0.970 & 0.766 & 0.700 \\
    A2S2K-ResNet \cite{roy2020attention} & 0.987 & 0.985 & 0.995 & \textbf{0.998} & \textbf{0.993} & \textbf{0.993} & 0.809 & 0.757 \\
    FreeNet \cite{zheng2020fpga} & - & - & \textbf{0.998} & 0.997 & - & - & \textbf{0.847} & \textbf{0.806} \\
    \midrule
    PyResNet \cite{paoletti2018deep} & 0.915 & 0.902 & 0.971 & 0.961 & 0.959 & 0.954 & 0.768 & 0.722 \\
    PyResNet-MSST & \textbf{0.989} & \textbf{0.985} & 0.994 & 0.984 & 0.975 & 0.972 & 0.835 & 0.790 \\
    \bottomrule
  \end{tabular}
  \label{tab:MSSA}
\end{table*}

\begin{figure}[t]
  \centering
  \includegraphics[width=1\linewidth]{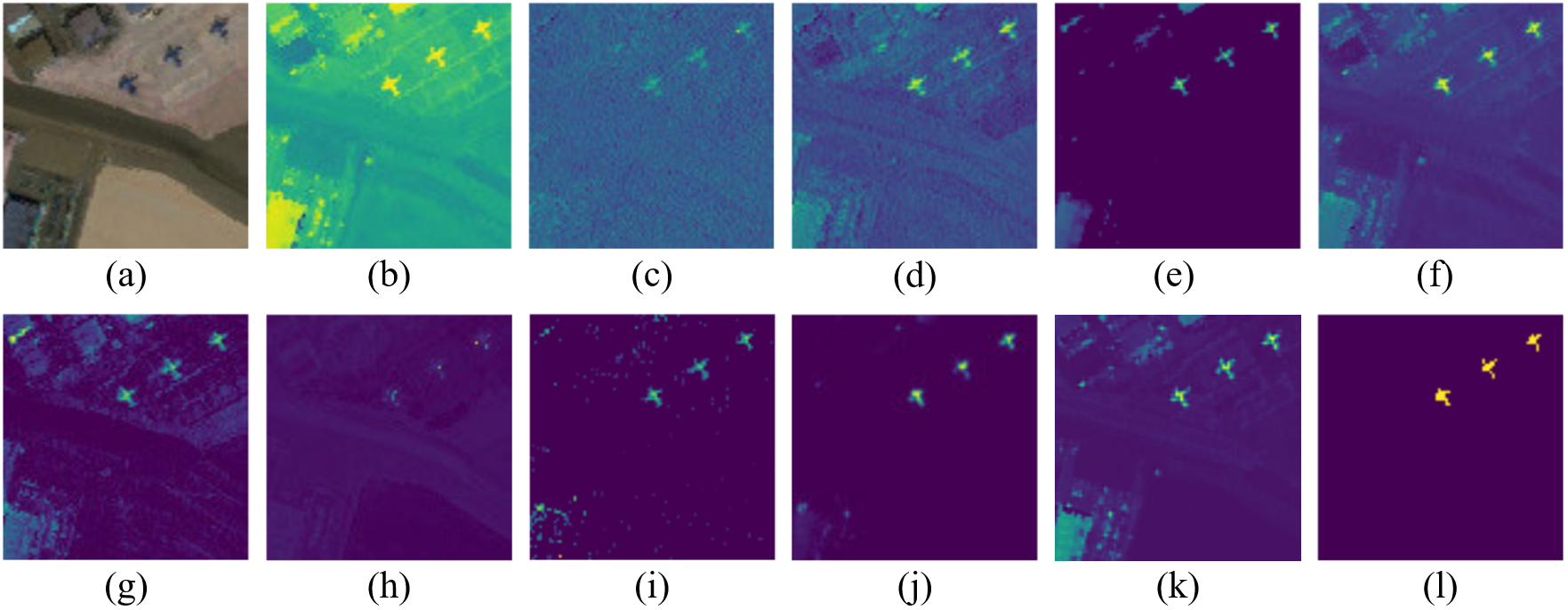}
  \caption{Visual target detection results of multiple methods on the San Diego dataset. (a) Pseudo-color image. (b) SAM. (c) CEM. (d) hCEM. (e) E-CEM. (f) CSCR. (g) DM-BDL. (h) HTD-Net. (i) BLTSC. (j) HTD-IRN. (k) DMSSN(ours). (l) Ground truth.}
  \label{fig:RSTD}
\end{figure}

\subsection{Extension on Remote Sensing Images}
The proposed Distilled Mixed Spectral-Spatial Network (DMSSN) not only demonstrates state-of-the-art capabilities in hyperspectral salient target detection but also holds significant potential for various other hyperspectral computer vision tasks. Given the widespread application of hyperspectral image processing technology in remote sensing, we conducted experiments to evaluate the performance of DMSSN in remote sensing target detection and classification tasks.

In the remote sensing target detection task, we conducted experiments using the San Diego dataset for comparative analysis. This dataset comprises 189 spectral bands and focuses on detecting airplanes as the target objects \cite{wang2022meta}. The objective was to assess the performance of the proposed DMSSN and its variants, examining the effectiveness of each module. Additionally, we compared our DMSSN against several existing state-of-the-art methods, including SAM \cite{kruse1993spectral}, CEM \cite{farrand1997mapping}, hCEM \cite{zou2015hierarchical}, E-CEM \cite{zhao2019ensemble}, CSCE \cite{li2015combined}, DM-BDL \cite{cheng2021decomposition}, HTD-Net \cite{zhang2020htd}, BLTSC \cite{xie2020background} and HTD-IRN \cite{shen2023hyperspectral}. For a fair comparison, all the methods, including DMSSN and its variants, were trained using the same strategy as outlined in \cite{shen2023hyperspectral}. The quantitative analysis results are presented in \Cref{tab:RSTD}, and qualitative visual comparisons are illustrated in \Cref{fig:RSTD}.

To elaborate on the specific variants, ``DMSSN-p'' indicates that only the Mixed Spectral-Spatial Transformer (MSST) network is utilized for extracting features, while ``DMSSN-s'' signifies the introduction of the spectral homogenization process. The complete ``DMSSN'' includes both the Distilled Spectral Encoding process and the MSST network. Notably, the computational time slightly increases with the introduction of spectral homogenization, but this process proves effective in improving accuracy. The incorporation of the complete Distilled Spectral Encoding process achieves high-quality data dimension reduction, resulting in a substantial improvement in model accuracy and efficiency. The comparison between different variants of DMSSN strongly validates the effectiveness of each module. Additionally, it is evident that the proposed DMSSN strikes a balance between performance and efficiency when compared to existing methods. For instance, in comparison to the latest HTD-IRN method, DMSSN significantly improves the AUC with only a minor increase in computational time.  When compared to the highest-precision CSCR method, DMSSN exhibits improved computational efficiency by several orders of magnitude. In summary, DMSSN achieves state-of-the-art performance in remote sensing target detection tasks.

\begin{figure}[t]
  \centering
  \includegraphics[width=1\linewidth]{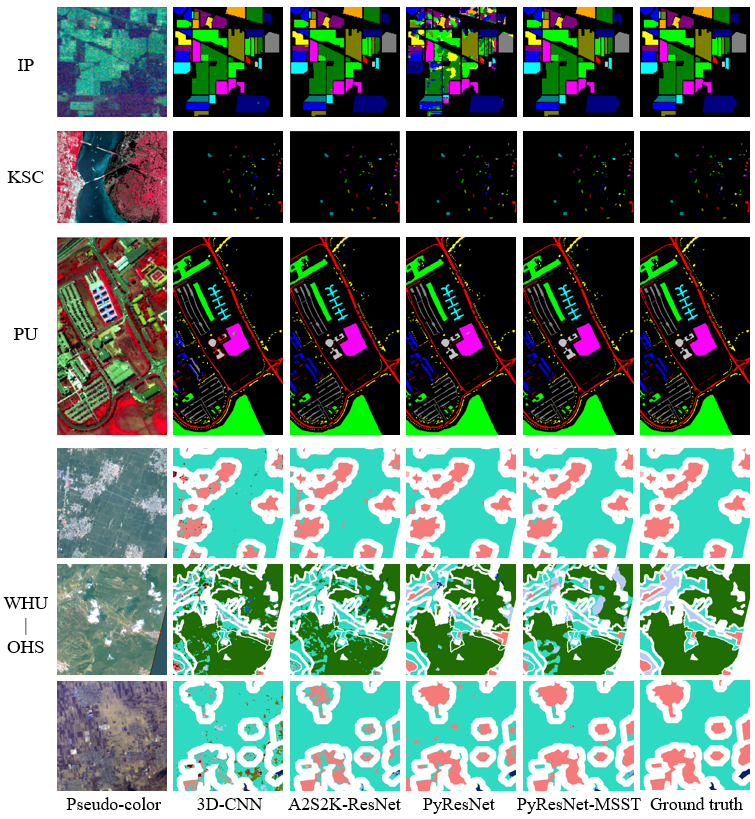}
  \caption{Visual classification results of multiple methods on IP, KSC, PU, and WHU-OHS datasets.}
  \label{fig:HSIC}
\end{figure}

Moreover, we have also validated the scalability of the proposed MSST feature extraction backbone network in remote sensing image classification tasks. Specifically, use PyResNet \cite{paoletti2018deep} as the basis, which consists of three parts: the input network consisting of convolutional and BatchNorm layers; the feature extraction backbone network comprising residual blocks; and the output network made up of pooling and fully connected layers. Considering the data format and task output requirements, we preserved the input and output network components. Subsequently, we introduced a novel method named PyResNet-MSST, where we replaced the feature extraction component with the proposed MSST. The training settings were kept consistent with the previous \cite{paoletti2018deep} to ensure fair comparison. The datasets employed include Indian Pines (IP) \cite{hong2021spectralformer}, Pavia Centre and University (PU) \cite{zou2022lessformer}, Kennedy Space Center (KSC) \cite{yan2021global}, and WHU-Orbita hyperspectral satellite (WHU-OHS) \cite{li2022whu}. 

We conduct a comprehensive comparison between PyResNet-MSST and several other methods, and we present the outcomes in \Cref{tab:MSSA}. Notably, the original PyResNet exhibits inferior performance compared to existing well-established methods across several datasets. For instance, considering the Indian Pines dataset, PyResNet displays a lower Overall Accuracy (OA) by $0.072$ and a lower Kappa coefficient by $0.083$ when compared to A2S2K-ResNet. However, with the integration of the MSST into PyResNet, the modified PyResNet-MSST achieves state-of-the-art performance. This augmentation led to a substantial enhancement in OA by $0.074$ and Kappa by $0.083$ on the Indian Pines dataset. Furthermore, the utilization of MSST as a feature extraction module yields significant improvements in classification accuracy across other datasets as well. For a visual representation of the results, \Cref{fig:HSIC} displays examples of visual classification results for multiple methods across the aforementioned four datasets. Notably, due to the multiple images within the WHU-OHS dataset, we have selected only a subset of the test set for visualization purposes. These visualizations collectively support the qualitative analysis that the MSST exhibits a robust feature extraction capability.

These experiments underscore the applicability of DMSSN in the remote sensing field, showcasing its competitiveness in tasks such as remote sensing target detection and classification. In fact, the proposed DMSSN is not limited to hyperspectral salient target detection but exhibits potential in various other hyperspectral computer vision applications.

\subsection{Ablation Study}

To evaluate the effectiveness of each component, we conducted multiple sets of ablation experiments. For a fair comparison, we maintain the same training settings as described in \Cref{subsec5.1}. Each innovative component is analyzed in detail below.

\begin{table*}[t]
\caption{Comparison of different data dimension reduction approaches. PCA: Principal Component Analysis. CONV: Convolution Layer. AE: Autoencoder. SH: 
Spectral Homogenization. KD: Knowledge Distillation.}
\centering
\setlength{\tabcolsep}{12pt}
  \begin{tabular}{ccccc|ccccc}
    \toprule
     PCA & CONV & AE & SH & KD & MAE $\downarrow$ & avg$\rm{F}_1$ $\uparrow$ & AUC $\uparrow$ & CC $\uparrow$ & NSS $\uparrow$ \\
    \midrule
    \checkmark & $\times$ & $\times$ & $\times$ & $\times$ & 0.133 & 0.197 & 0.618 & 0.205 & 0.595 \\
    $\times$ & \checkmark & $\times$ & $\times$ & $\times$ & 0.388 & 0.346 & 0.605 & 0.347 & 1.035 \\
    $\times$ & $\times$ & \checkmark & $\times$ & $\times$ & 0.125 & 0.505 & 0.750& 0.494 & 1.609 \\
    $\times$ & $\times$ & \checkmark & \checkmark & $\times$ & 0.117 & 0.527 & 0.766 & 0.522 & 1.710 \\
    \midrule
    $\times$ & $\times$ & \checkmark & \checkmark & \checkmark & \textbf{0.086}  & \textbf{0.621} & \textbf{0.817} & \textbf{0.636} & \textbf{2.171} \\
    \bottomrule
  \end{tabular}
  \label{tab:Autoencoder}
\end{table*}

\subsubsection{Effect of Autoencoder}
With the aid of knowledge distillation, the lightweight autoencoder is capable of achieving data dimension reduction while retaining sufficient information. To assess the effectiveness of this strategy, we conducted a series of comparative experiments, as illustrated in \Cref{tab:Autoencoder}. The baseline method, presented in the first row, entails data dimension reduction utilizing the traditional PCA method. The second row introduces an assessment of the impact of data dimension reduction through convolutional layers. Notably, we adopted the student autoencoder as part of the proposed DMSSN. The teacher autoencoder is used only for knowledge distillation.

The outcomes reveal a substantial enhancement in performance with the implementation of the student autoencoder, even in the absence of the knowledge distillation strategy. For instance, in comparison to the traditional PCA method, the pure student autoencoder exhibits superior detection performance, manifesting a noteworthy improvement of avg$\rm{F}_1$ by $0.141$. The convolutional layer, while providing dimensionality reduction, neglects spectral characteristics, resulting in increased MAE and inferior performance metrics compared to the autoencoder. Additionally, a comparison before and after the introduction of spectral homogenization exhibits notable improvement, particularly evident in the CC with an increment of $0.028$. The subsequent employment of the knowledge distillation strategy further amplifies method performance, with the NSS metric witnessing a notable increase of $0.461$. It is crucial to emphasize that the introduction of the knowledge distillation strategy incurs no additional computational and memory costs, a facet that will be explicated in greater detail in the ensuing discussion.

\subsubsection{Balance of Knowledge Distillation Strategy}
To further investigate the impact of the knowledge distillation strategy on the autoencoder, we conduct analytical experiments encompassing both the encoding and decoding stages. The detailed experimental results are delineated in \Cref{tab:distillation}. The nomenclature ``ORI'' denotes the original hyperspectral data prior to dimension reduction, while ``PCA'' represents the traditional principal component analysis method. To ensure a fair comparison, all autoencoders, including teacher autoencoder (TEA), pure student autoencoder (STU), and student autoencoder with knowledge distillation strategy (DIS), are trained with identical settings. Specifically, the initial learning rate is $0.003$, the batch size is $6$, and the number of training epochs is $20$.

\begin{table*}[t]
\caption{Ablation on Distilled Spectral Encoding. IE Represents the Information Entropy and SCC Represents the Spectral Correlation Coefficient with the Original Data.} 
  \centering
  \setlength{\tabcolsep}{10pt}
  \begin{tabular}{l|cc|cccc|cc}
    \toprule
    Method & IE $\uparrow$ & SCC $\uparrow$ & Speed & FLOPs & \#Param & Mem & MSE $\downarrow$ & MAE $\downarrow$ \\
    \midrule
    Original HSI  & \textbf{4.67} & - & - & - & - & - & - & -\\
    Principal Component Analysis & 2.14 & 0.04 & 1.7 FPS & - & - & - & - & -\\
    Teacher Autoencoder & 3.52 & \textbf{0.17} & 3.3 FPS & 65.3 G & 256 M & 7.53 G & \textbf{0.06} & \textbf{0.15} \\
    Student Autoencoder & 1.85 & 0.07 & \textbf{10 FPS} & \textbf{8.07 G} & \textbf{31 M} & \textbf{3.73 G} & 0.15 & 0.20 \\
    \midrule
    Distilled Student Autoencoder & 3.16 & 0.08 & \textbf{10 FPS} & \textbf{8.07 G} & \textbf{31 M} & 3.96 G & 0.11 & 0.18 \\
   \bottomrule
  \end{tabular}
  \label{tab:distillation}
\end{table*}

\textbf{The Encoding Stage.}
For a comprehensive evaluation, we utilize several metrics to assess the encoding results of the above data dimension reduction methods. These metrics include Information Entropy (IE) and Spectral Correlation Coefficient with the original data (SCC) for encoding performance. Additionally, we quantify the computational complexity through the number of images processed per second (Speed), floating point operations (FLOPs), model parameters (\#Param), and memory consumption (Mem) of the three autoencoders, all trained under identical settings. 

Given the inevitable information loss associated with data dimension reduction, the original hyperspectral data without encoding contains the maximal content, reflected in the highest IE. Among the methods considered, PCA exhibits the least efficiency and experiences the most substantial information loss, resulting in the lowest correlation between the encoding result and the original data. The teacher autoencoder, owing to its deep network structure, achieves the closest IE to the original data and attains the highest SCC. However, this comes at the expense of computational and memory costs, undermining the primary intent of the Distilled Spectral Encoding process.

In contrast, the pure student autoencoder combines high efficiency with a compact model size, but its encoding ability falls noticeably short of that achieved by the teacher autoencoder. Importantly, the knowledge distillation strategy, without altering the network structure, incurs no additional cost compared to the pure student autoencoder. Notably, the distilled student autoencoder exhibits robust encoding ability, a lightweight model size, and high efficiency.

\textbf{The Decoding Stage.}
As the decoding stage of the autoencoder solely requires the encoded feature map as input, the reconstructed similarity objectively reflects the performance of dimension reduction. \Cref{tab:distillation} presents the similarity between the data reconstruction outcomes of the autoencoders and the original hyperspectral data, measured through Mean Square Error (MSE) and Mean Absolute Error (MAE), where a smaller value indicates higher similarity.

The deep network structure enhances the encoding-decoding ability of the teacher autoencoder, resulting in the lowest MAE and MSE. However, due to substantial increases in computation and memory costs, the teacher autoencoder becomes impractical for practical use. In contrast, the pure student autoencoder, while reducing computational costs, exhibits the lowest MAE and MSE simultaneously.

To address this trade-off, we construct the distilled student autoencoder with the knowledge distillation strategy, enhancing the encoding-decoding ability with decreased MAE of $0.04$ and MSE of $0.02$. Both student autoencoders share the same structure, resulting in consistent FLOPs and the number of parameters. Loading the pre-trained model of the teacher autoencoder causes a slight increase in memory consumption. These results affirm that the Distilled Spectral Encoding process yields a lightweight autoencoder with robust encoding ability and high efficiency.

\begin{table*}[t]
\caption{Comparative results of different network configurations. Feature extraction represents the different modules for processing the data generated by the autoencoder. Scale fusion indicates different levels of feature utilization.}
\centering
\setlength{\tabcolsep}{12pt}
  \begin{tabular}{c|c|ccccc}
    \toprule
     Feature Extraction & Scale Fusion & MAE $\downarrow$ & avg$\rm{F}_1$ $\uparrow$ & AUC $\uparrow$ & CC $\uparrow$ & NSS $\uparrow$ \\
    \midrule
    ViT & Multiple & 0.115 & 0.523 & 0.735 & 0.516 & 1.845 \\
    CSP & Multiple & 0.100 & 0.562 & 0.795 & 0.588 & 1.912 \\
    MSS* & Multiple & 0.108 & 0.554 & 0.750 & 0.553 & 1.854 \\
    MSS & Single$_u$ & 0.244 & 0.468 & 0.660 & 0.449 & 1.169 \\
    MSS & Single$_b$ & 0.104 & 0.594 & 0.792 & 0.566 & 1.689 \\
    MSS & Double$_{ub}$ & 0.426 & 0.378 & 0.622 & 0.375 & 0.999 \\
    \midrule
    MSS & Multiple & \textbf{0.086}  & \textbf{0.621} & \textbf{0.817} & \textbf{0.636} & \textbf{2.171} \\
    \bottomrule
  \end{tabular}
  \label{tab:mssblock}
\end{table*}

\subsubsection{Effect of Mixed Spectral-Spatial Block}
To evaluate the effectiveness of the proposed Mixed Spectral-Spatial (MSS) block, we conducted a comprehensive comparison across various configurations, as detailed in \Cref{tab:mssblock}. Initially, we compared four feature extraction modules: the conventional self-attention mechanism in ViT \Cref{tab:mssblock} (ViT), the convolutional layers in YOLOv5 \cite{du2023sarnas} (CSP), the proposed MSS block (MSS), and its variant (MSS*). Results indicate that the MSS block outperforms conventional self-attention by fully leveraging hyperspectral imaging (HSI) characteristics through the employed spectral-spatial multi-head attention. Compared with the advanced convolution-based module, MSS remains competitive. Moreover, we evaluate the performance of a MSS variant, as shown in the third row. The original MSS block obtains ${\textbf{\textit{K}}}$ and ${\textbf{\textit{V}}}$ using different weighted convolutional layers, whereas the variant uses convolutional layers with shared weights to keep ${\textbf{\textit{K}}}$ and ${\textbf{\textit{V}}}$ equal. The MSS block demonstrates higher detection performance than the variant, as indicated by a significant increase in AUC of $0.067$, suggesting the necessity of separately acquiring ${\textbf{\textit{K}}}$ and ${\textbf{\textit{V}}}$.

The proposed MSS block generates pyramid-level features and fuses multi-scale features to predict the final detection result. As depicted in \Cref{tab:mssblock}, we conducted a comprehensive analysis comparing the performance of information fusion across varying scales. In our exploration, ``Single$_u$'' refers to utilizing only the uppermost feature, capturing primarily shallow spatial information with the largest scale. ``Single$_b$'' indicates the use of the bottommost feature with the smallest scale, emphasizing deep semantic information. ``Double$_{ub}$'' involves fusing the largest and smallest scale features, while ``Multiple'' represents the fusion of features across multiple scales. Our findings reveal that deep features outperform shallow features, attributed to their stronger object representation capabilities. Interestingly, dual-scale fusion proves less effective, as the large difference between two scale features lacks necessary transition information, thereby harming model performance. Consequently, the multi-scale fusion emerges as the optimal configuration, demonstrating superior performance across all metrics.



The results underscore that the proposed DMSSN, with all innovations enabled, outperforms other models in terms of salient object detection performance. The introduction of the autoencoder brings about a significant performance improvement, with a $70\%$ reduction in MAE and increases of $35\%$, $23\%$, $35\%$, and $64\%$ in avgF$_1$, AUC, CC and NSS, respectively. The knowledge distillation strategy further enhances the model performance without incurring additional computational costs, with avg$\rm{F}_1$ improving by $0.094$, and NSS improving by $0.461$. The adoption of MSS blocks effectively contributes to performance enhancement, as evidenced by an increase in AUC of $0.082$. In summary, each component fulfills its designated function, rendering the proposed DMSSN an effective solution for the HSOD task.

\section{Future Work}
While this study has yielded improvements in detection performance and dataset expansion, it faces remaining limitations that require future solutions.
Primarily, DMSSN's effectiveness hinges on abundant training samples, leading to underperformance in data-scarce scenarios like HS-SOD.
Moreover, while the MSST network showcases the superiority of HSIs processing, its applicability to RGB image-based HSOD tasks warrants further investigation.
Additionally, HSOD-BIT mainly emphasizes single-target data and lacks multi-target data.
Furthermore, the dataset volume, while expanded, still falls short in comparison to RGB-based datasets. 
To tackle these challenges, future endeavors will involve optimizing the method and conducting further validation on RGB image-based salient object detection datasets. Additionally, efforts will be directed toward refining data collection strategies, expanding the dataset, and incorporating a higher proportion of multi-objective data.

\section{Conclusion}
In conclusion, this paper presents a groundbreaking approach in hyperspectral salient object detection (HSOD) by introducing the Distilled Mixed Spectral-Spatial Network (DMSSN), which leverages a novel Distilled Spectral Encoding process and a Mixed Spectral-Spatial Transformer (MSST) backbone network. Our method addresses the critical challenges in HSOD by efficiently reducing the spectral dimension without losing valuable information and by concurrently utilizing spectral and spatial features to enhance detection performance. The creation of the HSOD-BIT dataset, with its large volume and diversity in complex scenarios, represents a significant contribution to HSOD research, providing a robust platform for training and evaluating deep learning methods. Our experimental results confirm the superior performance of the DMSSN over existing RGB-based methods, especially in challenging conditions, and highlight the potential of our MSST backbone network in broader remote sensing applications. This work not only sets a new benchmark in HSOD but also opens new avenues for future research and development in the field of hyperspectral image processing, promising enhanced performance in a wide range of practical applications.

\bibliographystyle{IEEEtran}
\bibliography{bibliography}

\begin{thebibliography}{10}
\providecommand{\url}[1]{#1}
\csname url@samestyle\endcsname
\providecommand{\newblock}{\relax}
\providecommand{\bibinfo}[2]{#2}
\providecommand{\BIBentrySTDinterwordspacing}{\spaceskip=0pt\relax}
\providecommand{\BIBentryALTinterwordstretchfactor}{4}
\providecommand{\BIBentryALTinterwordspacing}{\spaceskip=\fontdimen2\font plus
\BIBentryALTinterwordstretchfactor\fontdimen3\font minus \fontdimen4\font\relax}
\providecommand{\BIBforeignlanguage}[2]{{%
\expandafter\ifx\csname l@#1\endcsname\relax
\typeout{** WARNING: IEEEtran.bst: No hyphenation pattern has been}%
\typeout{** loaded for the language `#1'. Using the pattern for}%
\typeout{** the default language instead.}%
\else
\language=\csname l@#1\endcsname
\fi
#2}}
\providecommand{\BIBdecl}{\relax}
\BIBdecl

\bibitem{li2023lightweight}
G.~Li, Z.~Liu, X.~Zhang, and W.~Lin, ``Lightweight salient object detection in optical remote-sensing images via semantic matching and edge alignment,'' \emph{IEEE Transactions on Geoscience and Remote Sensing}, vol.~61, pp. 1--11, 2023.

\bibitem{nasrabadi2013hyperspectral}
N.~M. Nasrabadi, ``Hyperspectral target detection: An overview of current and future challenges,'' \emph{IEEE Signal Processing Magazine}, vol.~31, no.~1, pp. 34--44, 2013.

\bibitem{gao2019optical}
Z.~Gao, Y.~Zhao, L.~R. Khot, G.-A. Hoheisel, and Q.~Zhang, ``Optical sensing for early spring freeze related blueberry bud damage detection: Hyperspectral imaging for salient spectral wavelengths identification,'' \emph{Computers and Electronics in Agriculture}, vol. 167, p. 105025, 2019.

\bibitem{wang2022multiscale}
Z.~Wang, J.~Guo, C.~Zhang, and B.~Wang, ``Multiscale feature enhancement network for salient object detection in optical remote sensing images,'' \emph{IEEE Transactions on Geoscience and Remote Sensing}, vol.~60, pp. 1--19, 2022.

\bibitem{zheng2023boundary}
Q.~Zheng, L.~Zheng, Y.~Bai, H.~Liu, J.~Deng, and Y.~Li, ``Boundary-aware network with two-stage partial decoders for salient object detection in remote sensing images,'' \emph{IEEE Transactions on Geoscience and Remote Sensing}, vol.~61, pp. 1--13, 2023.

\bibitem{li2019distortion}
J.~Li, J.~Su, C.~Xia, and Y.~Tian, ``Distortion-adaptive salient object detection in 360 omnidirectional images,'' \emph{IEEE Journal of Selected Topics in Signal Processing}, vol.~14, no.~1, pp. 38--48, 2019.

\bibitem{wu2022hyperspectral}
Z.~Wu, H.~Su, X.~Tao, L.~Han, M.~E. Paoletti, J.~M. Haut, J.~Plaza, and A.~Plaza, ``Hyperspectral anomaly detection with relaxed collaborative representation,'' \emph{IEEE Transactions on Geoscience and Remote Sensing}, vol.~60, pp. 1--17, 2022.

\bibitem{zhang2018salient}
L.~Zhang, Y.~Zhang, H.~Yan, Y.~Gao, and W.~Wei, ``Salient object detection in hyperspectral imagery using multi-scale spectral-spatial gradient,'' \emph{Neurocomputing}, vol. 291, pp. 215--225, 2018.

\bibitem{cai2021hypergraph}
Y.~Cai, Z.~Zhang, Z.~Cai, X.~Liu, and X.~Jiang, ``Hypergraph-structured autoencoder for unsupervised and semisupervised classification of hyperspectral image,'' \emph{IEEE Geoscience and Remote Sensing Letters}, vol.~19, pp. 1--5, 2021.

\bibitem{yu2022unsupervised}
C.~Yu, S.~Zhou, M.~Song, B.~Gong, E.~Zhao, and C.-I. Chang, ``Unsupervised hyperspectral band selection via hybrid graph convolutional network,'' \emph{IEEE Transactions on Geoscience and Remote Sensing}, vol.~60, pp. 1--15, 2022.

\bibitem{zhou2019learning}
P.~Zhou, J.~Han, G.~Cheng, and B.~Zhang, ``Learning compact and discriminative stacked autoencoder for hyperspectral image classification,'' \emph{IEEE Transactions on Geoscience and Remote Sensing}, vol.~57, no.~7, pp. 4823--4833, 2019.

\bibitem{imamoglu2018hyperspectral}
N.~Imamoglu, Y.~Oishi, X.~Zhang, G.~Ding, Y.~Fang, T.~Kouyama, and R.~Nakamura, ``Hyperspectral image dataset for benchmarking on salient object detection,'' in \emph{2018 Tenth international conference on quality of multimedia experience (qoMEX)}.\hskip 1em plus 0.5em minus 0.4em\relax IEEE, 2018, pp. 1--3.

\bibitem{itti1998model}
L.~Itti, C.~Koch, and E.~Niebur, ``A model of saliency-based visual attention for rapid scene analysis,'' \emph{IEEE Transactions on pattern analysis and machine intelligence}, vol.~20, no.~11, pp. 1254--1259, 1998.

\bibitem{zhao2023learning}
S.~Zhao, Z.~Wen, Q.~Qi, K.-M. Lam, and J.~Shen, ``Learning fine-grained information with capsule-wise attention for salient object detection,'' \emph{IEEE Transactions on Multimedia}, 2023.

\bibitem{liang2013salient}
J.~Liang, J.~Zhou, X.~Bai, and Y.~Qian, ``Salient object detection in hyperspectral imagery,'' in \emph{2013 IEEE International conference on image processing}.\hskip 1em plus 0.5em minus 0.4em\relax IEEE, 2013, pp. 2393--2397.

\bibitem{gu2023orsi}
Y.~Gu, H.~Xu, Y.~Quan, W.~Chen, and J.~Zheng, ``Orsi salient object detection via bidimensional attention and full-stage semantic guidance,'' \emph{IEEE Transactions on Geoscience and Remote Sensing}, vol.~61, pp. 1--13, 2023.

\bibitem{li2019nested}
C.~Li, R.~Cong, J.~Hou, S.~Zhang, Y.~Qian, and S.~Kwong, ``Nested network with two-stream pyramid for salient object detection in optical remote sensing images,'' \emph{IEEE Transactions on Geoscience and Remote Sensing}, vol.~57, no.~11, pp. 9156--9166, 2019.

\bibitem{li2021multi}
G.~Li, Z.~Liu, W.~Lin, and H.~Ling, ``Multi-content complementation network for salient object detection in optical remote sensing images,'' \emph{IEEE Transactions on Geoscience and Remote Sensing}, vol.~60, pp. 1--13, 2021.

\bibitem{imamouglu2019salient}
N.~{\.I}mamo{\u{g}}lu, G.~Ding, Y.~Fang, A.~Kanezaki, T.~Kouyama, and R.~Nakamura, ``Salient object detection on hyperspectral images using features learned from unsupervised segmentation task,'' in \emph{ICASSP 2019-2019 IEEE International Conference on Acoustics, Speech and Signal Processing (ICASSP)}.\hskip 1em plus 0.5em minus 0.4em\relax IEEE, 2019, pp. 2192--2196.

\bibitem{huang2021salient}
C.~Huang, T.~Xu, Y.~Zhang, C.~Pan, J.~Hao, and X.~Li, ``Salient object detection on hyperspectral images in wireless network using cnn and saliency optimization,'' \emph{Ad Hoc Networks}, vol. 112, p. 102369, 2021.

\bibitem{yang2021cross}
K.~Yang, H.~Sun, C.~Zou, and X.~Lu, ``Cross-attention spectral--spatial network for hyperspectral image classification,'' \emph{IEEE Transactions on Geoscience and Remote Sensing}, vol.~60, pp. 1--14, 2021.

\bibitem{roy2019hybridsn}
S.~K. Roy, G.~Krishna, S.~R. Dubey, and B.~B. Chaudhuri, ``Hybridsn: Exploring 3-d--2-d cnn feature hierarchy for hyperspectral image classification,'' \emph{IEEE Geoscience and Remote Sensing Letters}, vol.~17, no.~2, pp. 277--281, 2019.

\bibitem{dosovitskiy2020image}
A.~Dosovitskiy, L.~Beyer, A.~Kolesnikov, D.~Weissenborn, X.~Zhai, T.~Unterthiner, M.~Dehghani, M.~Minderer, G.~Heigold, S.~Gelly \emph{et~al.}, ``An image is worth 16x16 words: Transformers for image recognition at scale,'' \emph{arXiv preprint arXiv:2010.11929}, 2020.

\bibitem{roy2023spectral}
S.~K. Roy, A.~Deria, C.~Shah, J.~M. Haut, Q.~Du, and A.~Plaza, ``Spectral--spatial morphological attention transformer for hyperspectral image classification,'' \emph{IEEE Transactions on Geoscience and Remote Sensing}, vol.~61, pp. 1--15, 2023.

\bibitem{kang2023self}
X.~Kang, B.~Deng, P.~Duan, X.~Wei, and S.~Li, ``Self-supervised spectral--spatial transformer network for hyperspectral oil spill mapping,'' \emph{IEEE Transactions on Geoscience and Remote Sensing}, vol.~61, pp. 1--10, 2023.

\bibitem{qin2023factorization}
H.~Qin, D.~Zhou, T.~Xu, Z.~Bian, and J.~Li, ``Factorization vision transformer: Modeling long-range dependency with local window cost,'' \emph{IEEE Transactions on Neural Networks and Learning Systems}, 2023.

\bibitem{he2021spatial}
X.~He, Y.~Chen, and Z.~Lin, ``Spatial-spectral transformer for hyperspectral image classification,'' \emph{Remote Sensing}, vol.~13, no.~3, p. 498, 2021.

\bibitem{cao2015salient}
Y.~Cao, J.~Zhang, Q.~Tian, L.~Zhuo, and Q.~Zhou, ``Salient target detection in hyperspectral images using spectral saliency,'' in \emph{2015 IEEE China Summit and international conference on signal and information processing (chinaSIP)}.\hskip 1em plus 0.5em minus 0.4em\relax IEEE, 2015, pp. 1086--1090.

\bibitem{yan2016salient}
H.~Yan, Y.~Zhang, W.~Wei, L.~Zhang, and Y.~Li, ``Salient object detection in hyperspectral imagery using spectral gradient contrast,'' in \emph{2016 IEEE International geoscience and remote sensing symposium (IGARSS)}.\hskip 1em plus 0.5em minus 0.4em\relax IEEE, 2016, pp. 1560--1563.

\bibitem{liu2022global}
Y.~Liu, S.~Zhang, Z.~Wang, B.~Zhao, and L.~Zou, ``Global perception network for salient object detection in remote sensing images,'' \emph{IEEE Transactions on Geoscience and Remote Sensing}, vol.~60, pp. 1--12, 2022.

\bibitem{wang2022hybrid}
Q.~Wang, Y.~Liu, Z.~Xiong, and Y.~Yuan, ``Hybrid feature aligned network for salient object detection in optical remote sensing imagery,'' \emph{IEEE Transactions on Geoscience and Remote Sensing}, vol.~60, pp. 1--15, 2022.

\bibitem{ren2020salient}
Q.~Ren, S.~Lu, J.~Zhang, and R.~Hu, ``Salient object detection by fusing local and global contexts,'' \emph{IEEE Transactions on Multimedia}, vol.~23, pp. 1442--1453, 2020.

\bibitem{liang2018material}
J.~Liang, J.~Zhou, L.~Tong, X.~Bai, and B.~Wang, ``Material based salient object detection from hyperspectral images,'' \emph{Pattern Recognition}, vol.~76, pp. 476--490, 2018.

\bibitem{zong2018deep}
B.~Zong, Q.~Song, M.~R. Min, W.~Cheng, C.~Lumezanu, D.~Cho, and H.~Chen, ``Deep autoencoding gaussian mixture model for unsupervised anomaly detection,'' in \emph{International conference on learning representations}, 2018.

\bibitem{sun2022spectral}
L.~Sun, G.~Zhao, Y.~Zheng, and Z.~Wu, ``Spectral--spatial feature tokenization transformer for hyperspectral image classification,'' \emph{IEEE Transactions on Geoscience and Remote Sensing}, vol.~60, pp. 1--14, 2022.

\bibitem{lin2017feature}
T.-Y. Lin, P.~Doll{\'a}r, R.~Girshick, K.~He, B.~Hariharan, and S.~Belongie, ``Feature pyramid networks for object detection,'' in \emph{Proceedings of the IEEE conference on computer vision and pattern recognition}, 2017, pp. 2117--2125.

\bibitem{wang2022pvt}
W.~Wang, E.~Xie, X.~Li, D.-P. Fan, K.~Song, D.~Liang, T.~Lu, P.~Luo, and L.~Shao, ``Pvt v2: Improved baselines with pyramid vision transformer,'' \emph{Computational Visual Media}, vol.~8, no.~3, pp. 415--424, 2022.

\bibitem{li2023delving}
J.~Li, W.~Ji, M.~Zhang, Y.~Piao, H.~Lu, and L.~Cheng, ``Delving into calibrated depth for accurate rgb-d salient object detection,'' \emph{International Journal of Computer Vision}, vol. 131, no.~4, pp. 855--876, 2023.

\bibitem{wang2017salient}
J.~Wang, H.~Jiang, Z.~Yuan, M.-M. Cheng, X.~Hu, and N.~Zheng, ``Salient object detection: A discriminative regional feature integration approach.'' \emph{International Journal of Computer Vision}, vol. 123, no.~2, 2017.

\bibitem{wu2022edn}
Y.-H. Wu, Y.~Liu, L.~Zhang, M.-M. Cheng, and B.~Ren, ``Edn: Salient object detection via extremely-downsampled network,'' \emph{IEEE Transactions on Image Processing}, vol.~31, pp. 3125--3136, 2022.

\bibitem{qin2019basnet}
X.~Qin, Z.~Zhang, C.~Huang, C.~Gao, M.~Dehghan, and M.~Jagersand, ``Basnet: Boundary-aware salient object detection,'' in \emph{Proceedings of the IEEE/CVF conference on computer vision and pattern recognition}, 2019, pp. 7479--7489.

\bibitem{qin2020u2}
X.~Qin, Z.~Zhang, C.~Huang, M.~Dehghan, O.~R. Zaiane, and M.~Jagersand, ``U2-net: Going deeper with nested u-structure for salient object detection,'' \emph{Pattern recognition}, vol. 106, p. 107404, 2020.

\bibitem{zhuge2022salient}
M.~Zhuge, D.-P. Fan, N.~Liu, D.~Zhang, D.~Xu, and L.~Shao, ``Salient object detection via integrity learning,'' \emph{IEEE Transactions on Pattern Analysis and Machine Intelligence}, 2022.

\bibitem{gao2022weakly}
S.~Gao, W.~Zhang, Y.~Wang, Q.~Guo, C.~Zhang, Y.~He, and W.~Zhang, ``Weakly-supervised salient object detection using point supervison,'' in \emph{Proceedings of the AAAI Conference on Artificial Intelligence}, vol.~36, no.~1, 2022, pp. 670--678.

\bibitem{kruse1993spectral}
F.~A. Kruse, A.~Lefkoff, J.~Boardman, K.~Heidebrecht, A.~Shapiro, P.~Barloon, and A.~Goetz, ``The spectral image processing system (sips)—interactive visualization and analysis of imaging spectrometer data,'' \emph{Remote sensing of environment}, vol.~44, no. 2-3, pp. 145--163, 1993.

\bibitem{farrand1997mapping}
W.~H. Farrand and J.~C. Harsanyi, ``Mapping the distribution of mine tailings in the coeur d'alene river valley, idaho, through the use of a constrained energy minimization technique,'' \emph{Remote Sensing of Environment}, vol.~59, no.~1, pp. 64--76, 1997.

\bibitem{zou2015hierarchical}
Z.~Zou and Z.~Shi, ``Hierarchical suppression method for hyperspectral target detection,'' \emph{IEEE transactions on geoscience and remote sensing}, vol.~54, no.~1, pp. 330--342, 2015.

\bibitem{zhao2019ensemble}
R.~Zhao, Z.~Shi, Z.~Zou, and Z.~Zhang, ``Ensemble-based cascaded constrained energy minimization for hyperspectral target detection,'' \emph{Remote Sensing}, vol.~11, no.~11, p. 1310, 2019.

\bibitem{li2015combined}
W.~Li, Q.~Du, and B.~Zhang, ``Combined sparse and collaborative representation for hyperspectral target detection,'' \emph{Pattern Recognition}, vol.~48, no.~12, pp. 3904--3916, 2015.

\bibitem{cheng2021decomposition}
T.~Cheng and B.~Wang, ``Decomposition model with background dictionary learning for hyperspectral target detection,'' \emph{IEEE Journal of Selected Topics in Applied Earth Observations and Remote Sensing}, vol.~14, pp. 1872--1884, 2021.

\bibitem{zhang2020htd}
G.~Zhang, S.~Zhao, W.~Li, Q.~Du, Q.~Ran, and R.~Tao, ``Htd-net: A deep convolutional neural network for target detection in hyperspectral imagery,'' \emph{Remote Sensing}, vol.~12, no.~9, p. 1489, 2020.

\bibitem{xie2020background}
W.~Xie, X.~Zhang, Y.~Li, K.~Wang, and Q.~Du, ``Background learning based on target suppression constraint for hyperspectral target detection,'' \emph{IEEE Journal of Selected Topics in Applied Earth Observations and Remote Sensing}, vol.~13, pp. 5887--5897, 2020.

\bibitem{shen2023hyperspectral}
D.~Shen, X.~Ma, W.~Kong, J.~Liu, J.~Wang, and H.~Wang, ``Hyperspectral target detection based on interpretable representation network,'' \emph{IEEE Transactions on Geoscience and Remote Sensing}, 2023.

\bibitem{hong2021spectralformer}
D.~Hong, Z.~Han, J.~Yao, L.~Gao, B.~Zhang, A.~Plaza, and J.~Chanussot, ``Spectralformer: Rethinking hyperspectral image classification with transformers,'' \emph{IEEE Transactions on Geoscience and Remote Sensing}, vol.~60, pp. 1--15, 2021.

\bibitem{zou2022lessformer}
J.~Zou, W.~He, and H.~Zhang, ``Lessformer: Local-enhanced spectral-spatial transformer for hyperspectral image classification,'' \emph{IEEE Transactions on Geoscience and Remote Sensing}, vol.~60, pp. 1--16, 2022.

\bibitem{yan2021global}
P.~Yan, H.~Qin, J.~Wang, T.~Xu, L.~Song, H.~Li, and J.~Li, ``Global-local channel attention for hyperspectral image classification,'' in \emph{2021 International Conference on Electrical, Computer and Energy Technologies (ICECET)}.\hskip 1em plus 0.5em minus 0.4em\relax IEEE, 2021, pp. 1--6.

\bibitem{li2022whu}
J.~Li, X.~Huang, and L.~Tu, ``Whu-ohs: A benchmark dataset for large-scale hersepctral image classification,'' \emph{International Journal of Applied Earth Observation and Geoinformation}, vol. 113, p. 103022, 2022.

\bibitem{hu2015deep}
W.~Hu, Y.~Huang, L.~Wei, F.~Zhang, and H.~Li, ``Deep convolutional neural networks for hyperspectral image classification,'' \emph{Journal of Sensors}, vol. 2015, pp. 1--12, 2015.

\bibitem{li2017spectral}
Y.~Li, H.~Zhang, and Q.~Shen, ``Spectral--spatial classification of hyperspectral imagery with 3d convolutional neural network,'' \emph{Remote Sensing}, vol.~9, no.~1, p.~67, 2017.

\bibitem{roy2020attention}
S.~K. Roy, S.~Manna, T.~Song, and L.~Bruzzone, ``Attention-based adaptive spectral--spatial kernel resnet for hyperspectral image classification,'' \emph{IEEE Transactions on Geoscience and Remote Sensing}, vol.~59, no.~9, pp. 7831--7843, 2020.

\bibitem{zheng2020fpga}
Z.~Zheng, Y.~Zhong, A.~Ma, and L.~Zhang, ``Fpga: Fast patch-free global learning framework for fully end-to-end hyperspectral image classification,'' \emph{IEEE Transactions on Geoscience and Remote Sensing}, vol.~58, no.~8, pp. 5612--5626, 2020.

\bibitem{paoletti2018deep}
M.~E. Paoletti, J.~M. Haut, R.~Fernandez-Beltran, J.~Plaza, A.~J. Plaza, and F.~Pla, ``Deep pyramidal residual networks for spectral--spatial hyperspectral image classification,'' \emph{IEEE Transactions on Geoscience and Remote Sensing}, vol.~57, no.~2, pp. 740--754, 2018.

\bibitem{wang2022meta}
Y.~Wang, X.~Chen, F.~Wang, M.~Song, and C.~Yu, ``Meta-learning based hyperspectral target detection using siamese network,'' \emph{IEEE Transactions on Geoscience and Remote Sensing}, vol.~60, pp. 1--13, 2022.

\bibitem{du2023sarnas}
W.~Du, J.~Chen, C.~Zhang, P.~Zhao, H.~Wan, Z.~Zhou, Y.~Cao, Z.~Huang, Y.~Li, and B.~Wu, ``Sarnas: A hardware-aware sar target detection algorithm via multiobjective neural architecture search,'' \emph{IEEE Transactions on Geoscience and Remote Sensing}, 2023.

\end{thebibliography}

\newpage

\vspace{11pt}

\begin{IEEEbiography}
[{\includegraphics[width=1in,height=1.25in,clip,keepaspectratio]{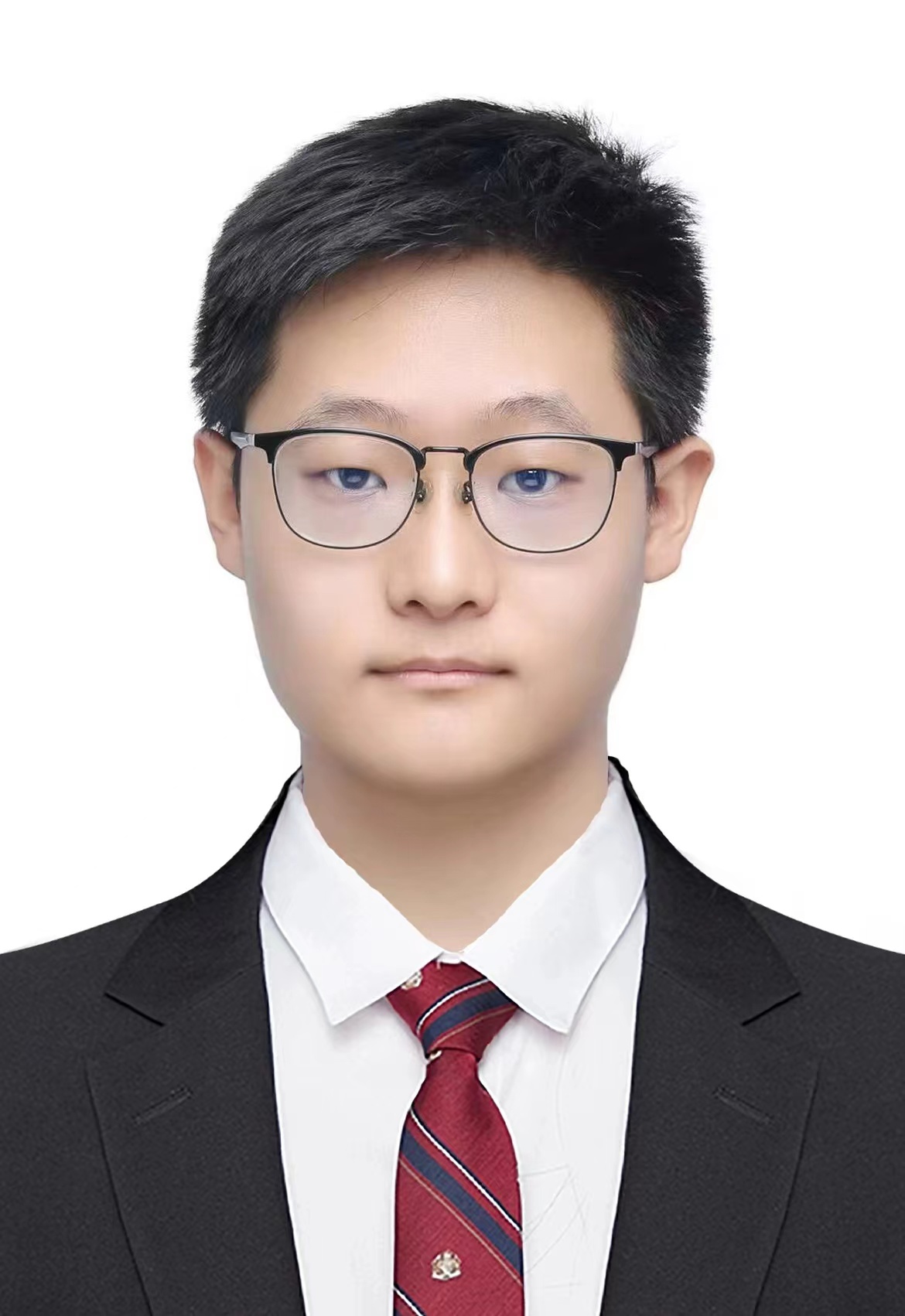}}]{Haolin Qin}
received the bachelor’s degree from the Beijing Institute of Technology,
Beijing, China, in 2020. He is currently
pursuing the Ph.D. degree in optical engineering with the Beijing Institute of Technology,
Beijing, China. His research interests
include computer vision, deep learning, and image processing.
\end{IEEEbiography}

\vspace{11pt}

\begin{IEEEbiography}[{\includegraphics[width=1in,height=1.25in,clip,keepaspectratio]{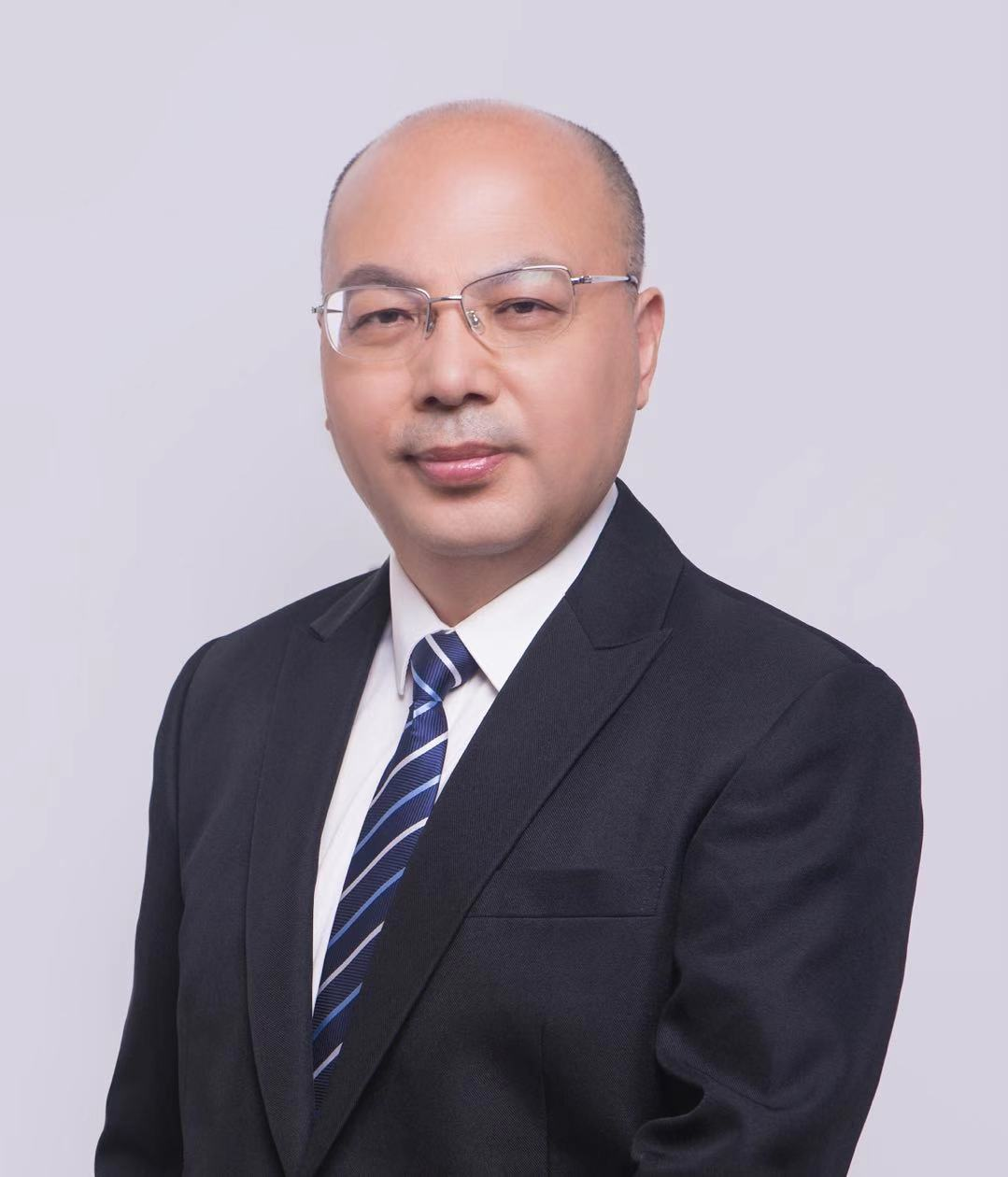}}]{Tingfa Xu}
received the Ph.D. degree from the Changchun Institute of Optics, Fine Mechanics and Physics, Changchun, China, in 2004. He is currently a Professor with the School of Optoelectronics, Beijing Institute of Technology, Beijing, China. His research interests include computer vision and optoelectronic imaging techniques.
\end{IEEEbiography}

\vspace{11pt}

\begin{IEEEbiography}[{\includegraphics[width=1in,height=1.25in,clip,keepaspectratio]{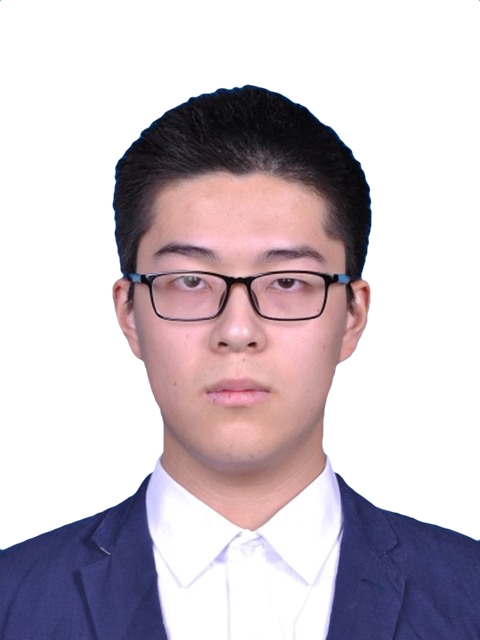}}]{Peifu Liu}
received his B.S. degree from the Beijing Institute of Technology, in 2021. He is currently pursuing his M.S. degree at the Beijing Institute of Technology, China. His current research interests include hyperspectral image processing and deep learning.
\end{IEEEbiography}

\vspace{11pt}

\begin{IEEEbiography}[{\includegraphics[width=1in,height=1.25in,clip,keepaspectratio]{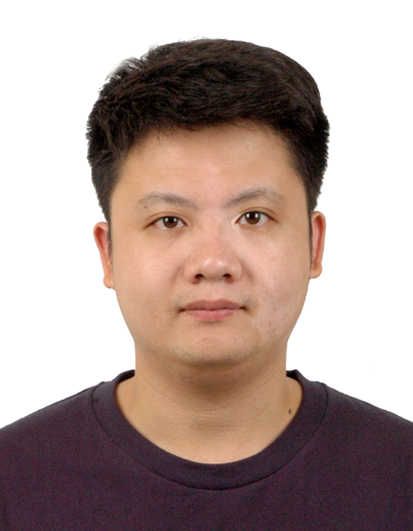}}]{Jingxuan Xu}
obtained his B.S. degree from Beijing University of Information Technology in 2021. He is pursuing his M.S. degree at Jiangsu Normal University - China Russia Academy. His current research interests include spectral analysis and signal modulation.
\end{IEEEbiography}

\vspace{11pt}

\begin{IEEEbiography}[{\includegraphics[width=1in,height=1.25in,clip,keepaspectratio]{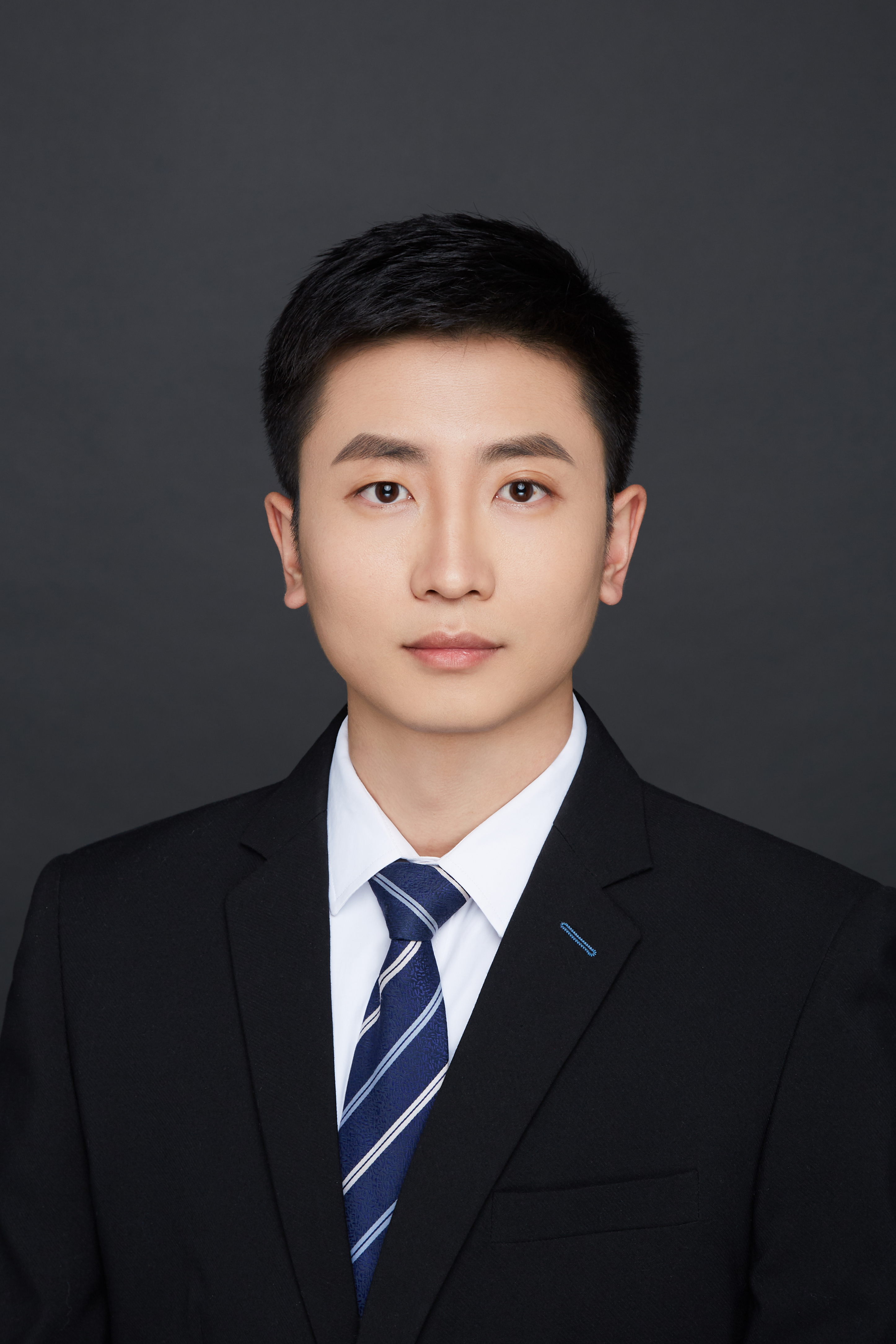}}]{Jianan Li}
is currently an Assistant Professor at School of Optoelectronics, Beijing Institute of Technology, Beijing, China, where he received his B.S. and Ph.D. degree in 2013 and 2019, respectively. From July 2019 to July 2020, he worked as a research fellow at National University of Singapore, where he also worked as a joint training Ph.D. student from July 2015 to July 2017. From October 2017 to April 2018, he worked as an intern at Adobe Research. His research interests mainly include computer vision and real-time image/video processing.
\end{IEEEbiography}

\vfill

\end{document}